%% file: SIGMOD-sigconf.tex
\documentclass[sigconf,10pt]{acmart}

\usepackage{graphicx}
\usepackage{balance} 
\usepackage{makecell}
\usepackage{xcolor}
\definecolor{green}{RGB}{0,102,51} 
\definecolor{orange}{RGB}{255,128,0}
\definecolor{velvet}{RGB}{138,43,226}
\definecolor{pink}{RGB}{219,112,147}
\usepackage{pgfplots}
\usepackage[inline]{enumitem}
\usepackage{tasks}
\usepackage{pgf}
\usepackage{subcaption}
\usepackage{url}
\usepackage{soul}
\usepackage{array}
\newcolumntype{P}[1]{>{\centering\arraybackslash}p{#1}}
\usepackage{multirow}
\usepackage{comment}

\def\BibTeX{{\rm B\kern-.05em{\sc i\kern-.025em b}\kern-.08emT\kern-.1667em\lower.7ex\hbox{E}\kern-.125emX}}
    
\copyrightyear{2020}
\acmYear{2020}
\setcopyright{acmcopyright}
\acmConference[SIGMOD'20]{Proceedings of the 2020 ACM SIGMOD International Conference on Management of Data}{June 14--19, 2020}{Portland, OR, USA}
\acmBooktitle{Proceedings of the 2020 ACM SIGMOD International Conference on Management of Data (SIGMOD'20), June 14--19, 2020, Portland, OR, USA}
\acmPrice{15.00}
\acmDOI{10.1145/3318464.3380599}
\acmISBN{978-1-4503-6735-6/20/06}

\input{chengkai-macros}

\input{macro.tex}

\begin{document}

\fancyhead{}

\title[Realistic Re-evaluation of Knowledge Graph Completion Methods]{Realistic Re-evaluation of Knowledge Graph Completion Methods: An Experimental Study}


\author{Farahnaz Akrami}
\email{farahnaz.akrami@mavs.uta.edu}
\affiliation{%
  \department{Department of Computer Science and Engineering}
  \institution{University of Texas at Arlington}
  }
\author{Mohammed Samiul Saeef}
\email{mohammedsamiul.saeef@mavs.uta.edu}
\affiliation{%
  \department{Department of Computer Science and Engineering}
  \institution{University of Texas at Arlington}
  }
\author{Qingheng Zhang}
\email{qhzhang.nju@gmail.com}
\affiliation{%
  \department{State Key Laboratory for Novel Software Technology}
  \institution{Nanjing University}
  }
\author{Wei Hu}
\email{whu@nju.edu.cn} 
\affiliation{%
  \department{State Key Laboratory for Novel Software Technology}
  \institution{Nanjing University}
  }
\author{Chengkai Li}
\email{cli@uta.edu}
\affiliation{%
  \department{Department of Computer Science and Engineering}
  \institution{University of Texas at Arlington}
  }
%
\renewcommand{\shortauthors}{Trovato and Tobin, et al.}

%
\input{sec-abstract}

\maketitle

\input{sec-introduction}
\input{sec-background}

\input{sec-eval}
\input{sec-issues}

\input{sec-exp}

\input{sec-conclusions}

%
\begin{acks}
Akrami, Saeef, and Li are partially supported by NSF grants IIS-1719054 and IIS-1937143. Zhang and Hu are partially supported by National Natural Science Foundation of China grant 61872172. Any opinions, findings, and conclusions or recommendations expressed in this publication are those of the authors and do not necessarily reflect the views of the funding agencies. We thank Lingbing Guo for his contribution to the preliminary report of this work. 

\end{acks}

%
\bibliographystyle{ACM-Reference-Format}
\bibliography{SIGMOD-sigconf}

\end{document}

%% file: chengkai-macros.tex
\newcommand{\eat}[1]{}
\newcommand{\gotoTR}[1]{}

\newcommand\xqed[1]{%
  \leavevmode\unskip\penalty9999 \hbox{}\nobreak\hfill
  \quad\hbox{#1}}

\renewcommand\qed{\xqed{\mbox{\raggedright \rule[0pt]{1.2ex}{1.2ex}}}}

\newcommand{\origunderscore}{}
\let\origunderscore\_
\renewcommand{\_}{\allowbreak\origunderscore}

%% file: macro.tex
\newcommand{\entity}[1]{\textsf{\small #1}}
\newcommand{\rel}[1]{\textsl{\small #1}}
\newcommand{\scriptsizerel}[1]{\textsl{\scriptsize #1}}
\newcommand{\triple}[3]{(\entity{#1}, \rel{#2}, \entity{#3})}

\newcommand{\hitone}{\texttt{\small Hits@1$^\uparrow$}}
\newcommand{\hone}{\texttt{\small H1$^\uparrow$}}
\newcommand{\hitten}{\texttt{\small Hits@10$^\uparrow$}}
\newcommand{\hten}{\texttt{\small H10$^\uparrow$}}
\newcommand{\mr}{\texttt{\small MR$^\downarrow$}}
\newcommand{\mrr}{\texttt{\small MRR$^\uparrow$}}
\newcommand{\fhitone}{\texttt{\small FHits@1$^\uparrow$}}
\newcommand{\fhone}{\texttt{\small FH1$^\uparrow$}}
\newcommand{\fhitten}{\texttt{\small FHits@10$^\uparrow$}}
\newcommand{\fhten}{\texttt{\small FH10$^\uparrow$}}
\newcommand{\fmr}{\texttt{\small FMR$^\downarrow$}}
\newcommand{\fmrr}{\texttt{\small FMRR$^\uparrow$}}

\newcommand{\tfhone}{\texttt{\tiny FH1$^\uparrow$}}

\newcommand{\tfhten}{\texttt{\tiny FH10$^\uparrow$}}
\newcommand{\tfmr}{\texttt{\tiny FMR$^\downarrow$}}
\newcommand{\tfmrr}{\texttt{\tiny FMRR$^\uparrow$}}

%% file: sec-abstract.tex
\begin{abstract}

In the active research area of employing embedding models for knowledge graph completion, particularly for the task of link prediction, most prior studies used two benchmark datasets FB15k and WN18 in evaluating such models. Most triples in these and other datasets in such studies belong to reverse and duplicate relations which exhibit high data redundancy due to semantic duplication, correlation or data incompleteness. This is a case of excessive data leakage---a model is trained using features that otherwise would not be available when the model needs to be applied for real prediction. There are also Cartesian product relations for which every triple formed by the Cartesian product of applicable subjects and objects is a true fact. Link prediction on the aforementioned relations is easy and can be achieved with even better accuracy using straightforward rules instead of sophisticated embedding models. 
A more fundamental defect of these models is that the link prediction scenario, given such data, is non-existent in the real-world.  This paper is the first systematic study with the main objective of assessing the true effectiveness of embedding models when the unrealistic triples are removed. Our experiment results show these models are much less accurate than what we used to perceive. Their poor accuracy renders link prediction a task without truly effective automated solution. Hence, we call for re-investigation of possible effective approaches.

\end{abstract}

%% file: sec-introduction.tex
\section{Introduction}\label{sec:intro}

Large-scale knowledge graphs such as Freebase~\cite{freebase}, DBpedia~\cite{DBpedia}, NELL~\cite{NELL}, Wikidata~\cite{wikidata}, and YAGO~\cite{yago} store real-world facts as triples in the form of \triple{head entity (subject)}{relation}{tail entity (object)}, denoted \triple{h}{r}{t}, e.g., \triple{Ludvig van Beethoven}{profession}{Composer}. They are an important resource for many AI applications, such as question answering~\cite{questionanswering,questionanswering-microsoft,questionanswering-rel}, search~\cite{eder2012knowledge}, and smart healthcare~\cite{healthkg}, to name just a few.  Despite their large sizes, knowledge graphs are far from complete in most cases, which hampers their usefulness in these applications. 

To address this important challenge, various methods have been proposed to automatically complete knowledge graphs. Existing methods in this active area of research can be categorized into two groups~\cite{nickel2016review}. One group is based on \emph{latent feature models}, also known as \emph{embedding models}, including TransE~\cite{bordes2013translating}, RESCAL~\cite{RESCAL}, and many other methods~\cite{cai2018comprehensive}.  
The other group is based on \emph{observed feature models} that exploit observable properties of a knowledge graph. Examples of such methods include rule mining systems~\cite{galarraga2013amie} and path ranking algorithms~\cite{lao2010relational}.

Particularly, the latent feature models are extensively studied. They embed each entity \entity{h} (or \entity{t}) into a multi-dimensional vector \textbf{h} (or \textbf{t}). A relation $r$ can have different representations. For example, in RESACL~\cite{RESCAL}, each relation is a weight matrix whose entries specify the interaction of latent features. In TransE~\cite{bordes2013translating}, a relation is a vector \textbf{r} that represents a geometric transformation between the head and tail entities in the embedding space and embeddings are learned in such a way that, if \triple{h}{r}{t} holds, then \(\textbf{h} + \textbf{r} \approx \textbf{t}\).

Embedding models have been extensively evaluated on \emph{link prediction}, a task that predicts the missing \entity{h} in triple \triple{?}{r}{t} or missing \entity{t} in \triple{h}{r}{?}. Two benchmark datasets FB15k (a subset of Freebase) and WN18 (extracted from WordNet~\cite{wordnet}), created by Bordes et al.~\cite{bordes2013translating}, are almost always used in such evaluation. Toutanova and Chen~\cite{toutanova2015observed} noted that FB15k contains many reverse triples, i.e., it includes many pairs of \triple{h}{r}{t} and \triple{t}{r$^{-1}$}{h} where \rel{r} and \rel{r}$^{-1}$ are reverse relations. They constructed another dataset, FB15k-237, by only keeping one relation out of any pair of reverse relations. Similarly, Dettmers et al.~\cite{ConvE} created WN18RR out of WN18 by removing reverse triples. The community has started to use FB15k-237 and WN18RR in evaluating models and noted significant performance degeneration of existing models in comparison with their performance on FB15k and WN18~\cite{ConvE,toutanova2015observed,kgcompletion}.

\begin{figure}
\small
\centering
\begin{subfigure}{.23\textwidth}
  \centering
  \includegraphics[trim={1cm 1.5cm 0cm 1.5cm},clip,width=.9\linewidth]{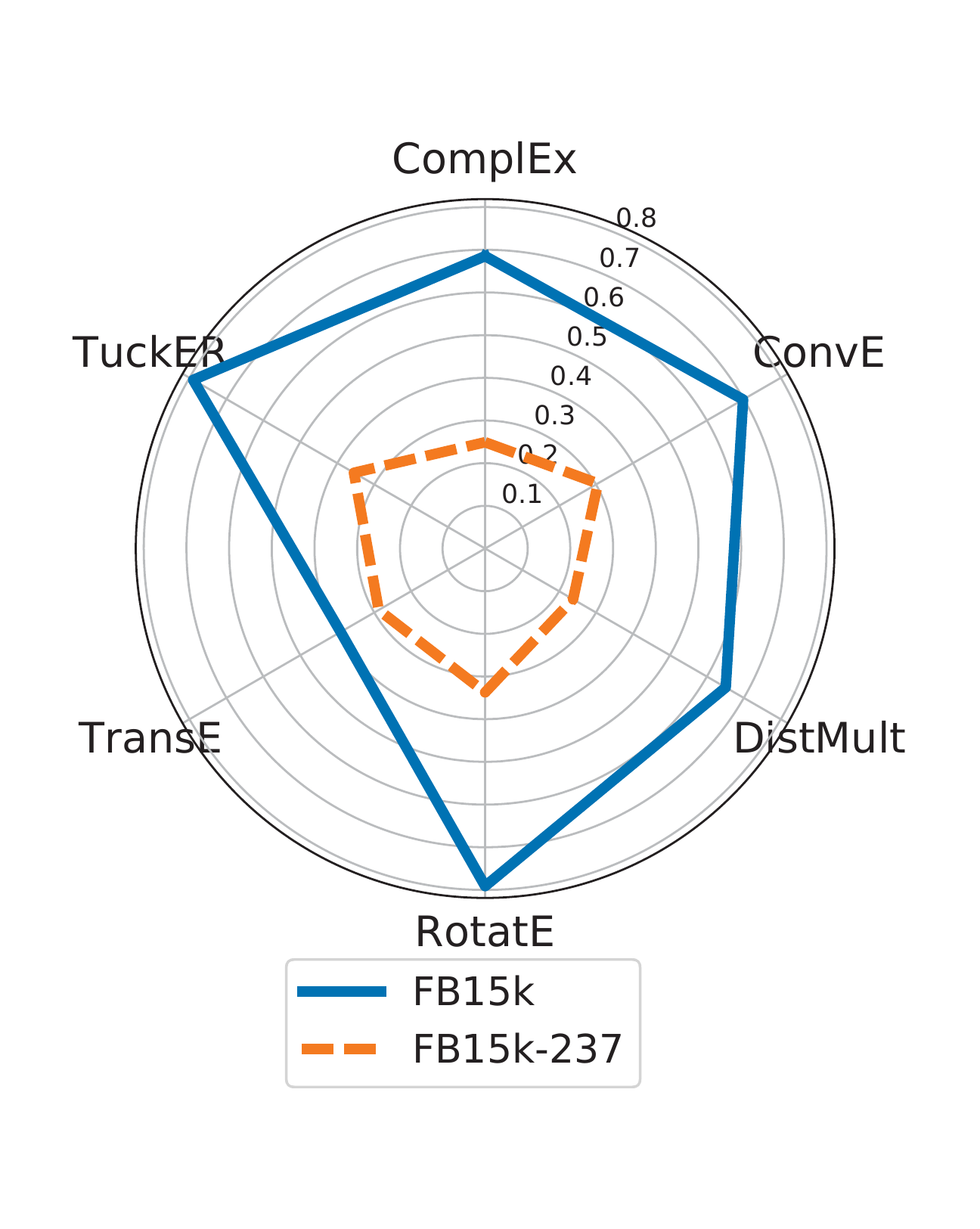}
  \label{fig:sub1}
\end{subfigure}\hspace{2mm}%
\begin{subfigure}{.23\textwidth}
  \centering
  \includegraphics[trim={0cm 1.5cm 1cm 1.5cm},clip,width=.9\linewidth]{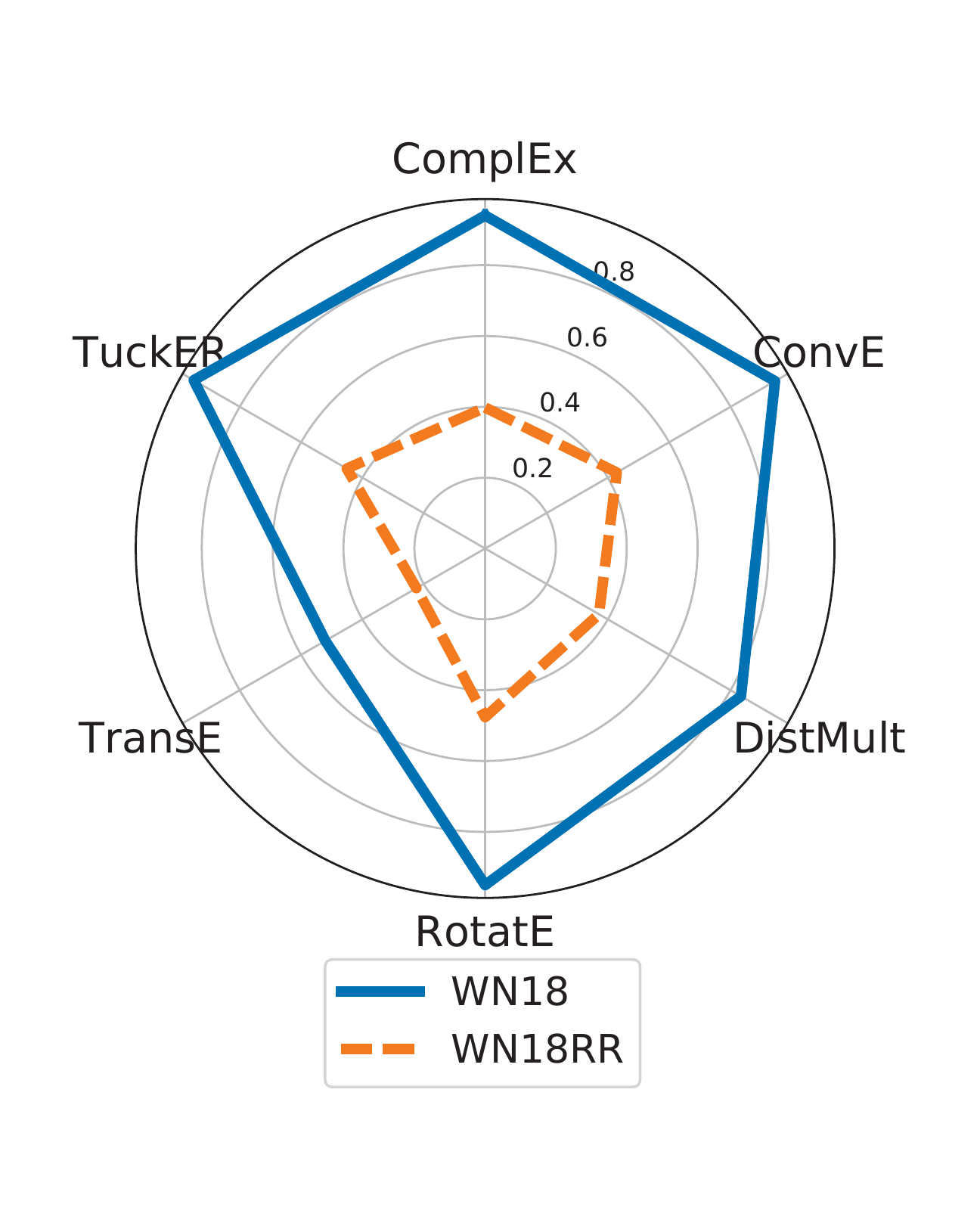}
  \label{fig:sub1}
\end{subfigure}%
\caption{\small Performance of embedding models on FB15k vs.~FB15k-237 and WN18 vs.~WN18RR using \fmrr}
\label{fig:radar}
\end{figure}

\textbf{Impact of reverse relations}:\hspace{2mm} This paper thoroughly examines the impact of reverse triples in FB15k and WN18 (details in Section~\ref{sec:reverserelation}). 
The idiosyncrasies of the link prediction task on such data can be summarized as follows. A1) \emph{Link prediction becomes much easier on a triple if its reverse triple is available.} A2) \emph{For reverse triples, a straightforward method could be even more effective than complex machine learning models.} We discovered that 70\% of the triples in the training set of FB15k form reverse pairs. Similarly, for 70\% of the triples in its test set, reverse triples exist in the training set. For WN18, these two percentages are even higher---92.5\% and 93\%.
The abundant reverse triples suggest that embedding models would have been biased toward learning whether two relations \rel{r}$_1$ and \rel{r$_2$} form a reverse pair. Instead of complex models, one may achieve this goal by using statistics of the triples to derive simple rules of the form \triple{h}{r$_1$}{t} $\Rightarrow$ \triple{t}{r$_2$}{h}. 
In fact, we generated such a simple model which attained 71.6\% for FB15k and 96.4\% for WN18 using \fhitone, a common accuracy measure for embedding models.~\footnote{An upward/downward arrow beside a measure indicates that methods with greater/smaller values by that measure possess higher accuracy.} These results are on par with those by the best performing embedding models---73.8\% and 94.6\% on FB15K and WN18, respectively, as can be seen from Table~\ref{table:Hits@1} in Section~\ref{sec:exp}.

The above analysis suggests that \emph{the reverse triples led to a substantial over-estimation of the embedding models' accuracy}, which is verified by our experiments on a wide range of models. While Section~\ref{sec:exp} examines the results in detail, Figure~\ref{fig:radar} illustrates the performance comparison of a few representative models using another popular measure \fmrr. The results show that R1) \emph{the performance of all existing embedding models degenerates significantly after reverse triples are removed}. 
R2) \emph{Many successors of the original TransE model  were empirically shown to outperform TransE by far on FB15k, but they only attained similar or even worse performance on FB15k-237}. For example, the \fhitten\ of ComplEX vs.~TransE is 42.3\% vs.~47.5\% on FB15k-237, in stark contrast to 83.2\% vs.~62.4\% on FB15k. R3) \emph{The absolute accuracy of all models is poor, rendering them ineffective for real-world link prediction task}. For example, TuckER~\cite{TuckER} attains the best \fmrr\ on FB15k-237 (0.355). However, its performance on FB15k (0.79) was considerably stronger. Similarly, RotatE~\cite{RotatE} has 0.95 \fmrr\ on WN18 but only 0.476 on WN18RR.

The existence of excessive reverse triples in FB15k and WN18---the de facto benchmark datasets for link prediction---actually presents a more fundamental defect in many of these models: A3) \emph{the link prediction scenario, given such data, is non-existent in the real-world at all}. With regard to FB15k, the redundant reverse relations, coming from Freebase, were just artificially created. When a new fact was added into Freebase, it would be added as a pair of reverse triples, denoted explicitly by a special relation \rel{reverse\_property}~\cite{freebasewikidata, semanticsearch}. 
In WN18, 17 out of the 18 relations are reverse relations. Some are reverse of each other, e.g., \rel{hypernym} and \rel{hyponym}---\entity{flower} is a hypernym of \entity{sunflower}  and \entity{sunflower} is a hyponym of \entity{flower}. Others are self-reciprocal, i.e., symmetric relations such as \rel{verb\_group}---\triple{begin}{verb\_group}{start} and \triple{start}{verb\_group}{begin} are both valid triples. 
For such intrinsically reverse relations that always come in pair when the triples are curated into the datasets, there is not a scenario in which one needs to predict a triple while its reverse is already in the knowledge graph. Training a knowledge graph completion model using FB15k and WN18 is thus a form of \emph{overfitting} in that the learned model is optimized for the reverse triples which cannot be generalized to realistic settings. More precisely, this is a case of excessive \emph{data leakage}---the model is trained using features that otherwise would not be available when the model needs to be applied for real prediction. There could be more natural reverse triples that are worth prediction---two relations are not semantically reverse but correlate and/or the reverse triples are not available together in the knowledge graph due to how the data are collected. We discuss such cases of data redundancy below.

\textbf{Impact of other data redundancy and Cartesian pro\-duct relations}:\hspace{2mm} The data leakage due to reverse triples is a form of data redundancy that unrealistically inflates the models' accuracy. We identified other types of data redundancy in FB15k and another evaluation dataset YAGO3-10 (Section~\ref{sec:duplicate}). Specifically, some relations are \emph{duplicate} as their subject-object pairs substantially overlap, and some are \emph{reverse duplicate} when one relation's subject-object pairs overlap a lot with another relation's object-subject pairs.

We also discovered another type of relations, which we call \emph{Cartesian product relations} (Section~\ref{sec:cartesian}), that unrealistically inflate a model's accuracy. Given such a relation, there are a set of subjects and a set of objects, and the relation is valid from every subject in the first set to every object in the second set. In a May 2013 snapshot of Freebase, close to 10\% of the relations are Cartesian product relations. In FB15k, 142 out of the 1345 relations are such relations. One example is \rel{position}, since every team in a certain professional sports league has the same set of positions. The link prediction problem for such relations thus becomes predicting, say, whether an NFL team has the quarter-back position, which is not very meaningful in the real-world. Moreover, when a substantial subset of the aforementioned subject-object Cartesian product is available in the training set, it is relatively easy for a model to attain a strong prediction accuracy. 

The same analyses A1-A3 on reverse relations are also applicable on duplicate and Cartersian product relations, and the same observations R1-R3 can be made from our experiment results. A1) In evaluating prediction models, it is misleading to mix such straightforward relations with more realistic, challenging relations. In the test set of FB15k, the numbers of reverse relations, duplicate and reverse duplicate relations, Cartesian product relations, and the remaining relations are 798, 118, 78, and 106, respectively. The \fmrr\ of ConvE on such relations is 0.72, 0.948, 0.881, and 0.444, respectively. Another example is YAGO3-10 which has two largely duplicate relations \rel{isAffiliatedTo} and \rel{playsFor} that account for more than 63\% of its test set. The \fmrr\ of RotatE~\cite{RotatE} is 0.612 on these 2 relations but only 0.304 on other relations. A2) Instead of learning complex embedding models, a simpler approach can be more effective. For duplicate and reverse duplicate relations, a simple rule based on data statistics can already be quite accurate, as similarly in the aforementioned case of reverse relations. For Cartesian product relations, by observing that a large percentage of possible subject-object pairs in a relation exist in the dataset, one can derive the relation is a Cartesian product relation and thus the same relation should exist in all such pairs. Our experiments on 9 Cartesian product relations in FB15k obtained an average \fhitten\ of 98.3\% using this method, which is higher than the 96.3\% \fhitten\ of TransE on these relations.
A3) The existence of Cartesian product relations in FB15k is quite artificial. In fact, 60\% of them are due to special ``mediator nodes'' in Freebase that represent multiary relationships and simplification in FB15k for removing such nodes through concatenating edges. 
Similarly, a vast majority of the duplicate and reverse duplicate relations in FB15k were artificially created. The dataset has 84 pairs of duplicate relations. In 80 out of the 84 pairs, one or both relations are concatenated. The numbers are 63 out of 67 pairs for reverse duplicate relations. Just like reverse triples, they render a link prediction scenario largely nonexistent in the real-world and lead to unrealistically strong prediction accuracy.

The much weaker performance of embedding models on FB15k-237 and WN18RR also drove us to examine observed feature models, specifically using rules discovered by the rule mining system AMIE~\cite{galarraga2013amie}. 
Our experiment results show that it also degenerates significantly on the more realistic FB15k-237 and WN18RR.
Its \fmrr\ on FB15k vs.~FB15k-237 is 0.797 vs.~0.308 and is 0.94 vs.~0.357 on WN18 vs.~WN18RR.

The embedding models generate a ranked list of candidate predictions which can be as long as the number of entities in a knowledge graph. For this ranked list to effectively assist human curators in completing the knowledge graph, the correct predictions should be ranked high. From this perspective, this study depicts a realistic picture of existing methods being much less accurate than one may perceive. As mentioned in R3, their absolute accuracy is poor, which renders link prediction a task without truly effective automated solution. Hence, we call for re-investigation of possible effective approaches for completing knowledge graphs.

This paper presents a systematic study with the main objective of assessing the true effectiveness of link prediction methods in real-world settings. Other studies continue to evaluate models using both FB15k and FB15k-237 (similarly WN18 and WN18RR), merely viewing the latter as a more challenging dataset. However, based on our analyses A1-A3 and experiment results R1-R3, \emph{we argue that FB15k and WN18 are completely misleading and should not be used anymore}. Similarly, our results show that YAGO3-10, which has been recently used in some studies~\cite{ConvE}, also suffers from the same defect since the majority of its triples are duplicates.

A preliminary report of our study was in~\cite{kgcompletion}, which pr\-es\-en\-ts much less comprehensive analyses and results. Other related studies are~\cite{wang-etal-2019-evaluating, finegrained}. The focus in \cite{wang-etal-2019-evaluating}, different from ours, is about how existing evaluation methods are more suitable for question answering than link prediction. Meilicke et al.~\cite{finegrained} analyzed the types of rules that help knowledge graph completion by categorizing test triples of datasets based on explanations generated by their rule-based approach. The test leakage problem was also mentioned in~\cite{ConvE}, although their focus was to propose a new embedding model.

To sum up, this paper makes these \textbf{contributions}:
\begin{list}{$\bullet$}
{ \setlength{\leftmargin}{1em} \setlength{\parsep}{0pt} \setlength{\itemsep}{0pt}}
    \item It provides a thorough investigation of the data redundancy problem in how existing embedding models for knowledge graph completion were trained, due to reverse and duplicate triples in the de facto benchmark datasets FB15k and WN18 (Section~\ref{sec:issues}).  
    \item For the first time, it identifies the existence of Cartesian product relations in FB15k which, together with the data redundancy problem, makes previous performance measures of embedding models unrealistic (Section~\ref{sec:issues}). 
    \item It presents the results of a comprehensive evaluation of these defects' impacts on the performance of many representative embedding models as well as  an observed feature model AMIE (Section~\ref{sec:exp}).
    \item All codes, experiment scripts, datasets, and results are in a public repository {\small \url{https://github.com/idirlab/kgcompletion}}. It will help ensure the reproducibility of this research. 
\end{list}

%% file: sec-background.tex
\section{Background: Knowledge Graph Completion Methods}\label{sec:background}

This section briefly summarizes representative knowledge graph completion methods. In our description, vectors are represented as bold lower case letters such as \textbf{x}. 
\([\textbf{x}]_i\) represents the $i$th element of \textbf{x}. A matrix is denoted by a bold upper case letter, e.g., \textbf{M}. 
A knowledge graph \(\mathcal{G}\) consists of a set of entities \(\mathcal{E}\) and a set of relations \(\mathcal{R}\). Triples are represented as \triple{h}{r}{t} where \entity{h}, \entity{t} \(\in  \mathcal{E}\) are the head and tail entities, and relationship \rel{r} \(\in \mathcal{R}\) exists from the head to the tail. \(\langle \textbf{x},\textbf{y},\textbf{z}\rangle=\sum_i [\textbf{x}]_i.[\textbf{y}]_i.[\textbf{z}]_i\) is the component-wise multi-linear dot product.  

\subsection{Latent Feature Models}\label{sec:embedding}

Embedding-based methods employ two crucial components: (1) a scoring function to measure the plausibility of triples \triple{h}{r}{t}, and (2) a process to learn the representations (i.e., embeddings) of entities and relations by solving an optimization problem of maximizing the scores of correct triples while minimizing the scores of incorrect ones.

In TransE~\cite{bordes2013translating}, the scoring function is $f_\rel{r} (\entity{h},\entity{t}) = -\Vert \textbf{h} + \textbf{r} - \textbf{t}\Vert_{\ell_1/\ell_2}^2$. 
TransE is a scalable method with a small number of model parameters, but it has limitations in modeling $1$-to-$n$, $n$-to-$1$, and $m$-to-$n$ relations \cite{TransH}. TransH~\cite{TransH} aims to address TransE's limitations by not using the same embedding of an entity in different relations. 

Lin et al.~\cite{TransR} proposed TransR which learns the embeddings in two different vector spaces $\mathbb{R}^d$ and $\mathbb{R}^k$ for entities and relations. They argued that using the same semantic space for entities and relations, as in TransE and TransH, is insufficient because they are two completely different types of objects. Instead, TransR defines a projection matrix \(\textbf{M}_r\) to map entity embeddings to the vector space for each relation. In TransD~\cite{TransD}, which improves over TransR, the projection matrix is decomposed to the product of two vectors. However, in contrast to TransR, there is a unique projection matrix for each entity-relation pair. RotatE~\cite{RotatE} defines each relation as a rotation from the source entity to the target entity. The scoring function is $f_\rel{r} (\entity{h},\entity{t}) = -{\Vert \textbf{h} \circ \textbf{r} -\textbf{t} \Vert}^2$, where \(\circ\) is the Hadmard (or element-wise) product.

To learn the entity and relation representations, a loss function is minimized. Two of the most-frequently used loss functions in embedding models are margin-based loss function $L= \sum_{\triple{h}{r}{t}\in S} \sum_{\triple{h'}{r}{t'}\in S'}max(0,f_\rel{r} (\entity{h},\entity{t}) + \gamma - f_\rel{r} (\entity{h'},\entity{t'}))$ and logistic loss $    L= \sum_{\triple{h}{r}{t}\in S\cup S'} log(1+exp(-y_{hrt}.f_\rel{r} (\entity{h},\entity{t})))$. 
In the equations, \(y_{hrt}\) is the sign of a training example (+1/-1 for positive/negative example), \(\gamma\) is the margin, \(S\) is the set of positive triples \triple{h}{r}{t} in the training set and \(S'\) is the set of negative triples \triple{h'}{r}{t'}. Since a real-world knowledge graph contains only positive triples, negative triples in evaluation datasets were generated by \emph{corrupting} the positive triples---a process that replaces the head or tail entity of each positive triple by other entities in the knowledge graph~\cite{bordes2013translating}. 

Another approach formulates link prediction as a third-order binary tensor completion problem in which a knowledge graph is represented as a partially observed tensor \(\textbf{Y}\in \{0,1\}^{|\mathcal{E}| \times |\mathcal{E}| \times |\mathcal{R}|}\). An entry in \(\textbf{Y}\) equals one if the corresponding triple exists in \(\mathcal{G}\). Different models such as RESCAL~\cite{RESCAL}, DistMult~\cite{DistMult}, ComplEx~\cite{complex}, and TuckER~\cite{TuckER} used various methods of tensor factorization to decompose \(\textbf{Y}\) and assign scores to triples based on the learned factors. 
RESCAL is a collective matrix factorization model which  represents a relation as a matrix $\textbf{W}_r\in \mathbb{R}^{d\times d}$ that describes the interactions between latent representations of entities. The score of a triple in this method is defined as $f_\rel{r} (\entity{h},\entity{t}) =  \textbf{h}^{\top}\textbf{W}_r \textbf{t}$. 
DistMult is similar to RESCAL but it restricts relations to be diagonal matrices \(\textbf{w}_r \in \mathbb{R}^d\) in order to reduce the number of relation parameters: $f_\rel{r} (\entity{h},\entity{t}) =  \langle\textbf{h},\textbf{w}_r,\textbf{t}\rangle$. 
Due to this simplification, DistMult can only model symmetric relations. ComplEx is an extension of DistMult. It uses complex numbers instead of real numbers to handle symmetric and anti-symmetric relations. 
TuckER is a model based on Tucker decomposition~\cite{tucker1964extension} of \(\textbf{Y}\). 
ConvE~\cite{ConvE} is a neural network model that uses 2D convolutional layers over embeddings, and interactions between entities and relations are modeled by convolutional and fully connected layers.

\vspace{-2mm}
\subsection{Other Approaches}\label{sec:amie}

Unlike embedding models which employ latent features, observed feature models directly exploit observable features. For instance, by observing that most persons in a knowledge graph have the same citizenship as their parents, a system may possibly generate rule \triple{a}{is\_citizen\_of}{c} $\wedge$ \triple{a}{has\_child}{b} $\Rightarrow$ \triple{b}{is\_citizen\_of}{c}, which can be also represented as  \rel{is\_citizen\_of}(\entity{a},\entity{c}) $\wedge$ \rel{has\_child}(\entity{a},\entity{b}) $\Rightarrow$ \rel{is\_citizen\_of}(\entity{b},\entity{c}). A representative of the observed feature models is the rule mining system AMIE~\cite{galarraga2013amie}.
In AMIE, a rule has a body (antecedent) and a head (consequent), represented as $B_1 \land B_2 \land \ldots \land B_n \Rightarrow H$
or in simplified form \(\overrightarrow{B} \Rightarrow H\).  
The body consists of multiple \textit{atoms} $B_1$, $\ldots$, $B_n$ and the head $H$ itself is also an atom. In an atom \rel{r}(\entity{h},\entity{t}), which is another representation of fact triple \triple{h}{r}{t}, the subject and/or the object are variables to be instantiated. The prediction of the head can be carried out when all the body atoms can be instantiated in the knowledge graph.

The multi-hop link prediction approaches~\cite{das2017go, multihop} aim to find complex patterns in a knowledge graph by following reasoning paths, e.g., $\entity{a}$$\to$$livesInCity$$\to$$\entity{b}$$\to$$isInCountry$$\to$$\entity{c}$.  Given a query \triple{$e$}{r}{?}, these approaches use reinforcement learning in learning to walk from \entity{$e$} to the answer entity by taking a labeled relation at each step, conditioned on the query relation and entire path history. 

%% file: sec-eval.tex
\section{Existing Evaluation Framework}\label{sec:eval}

\subsection{Evaluation Datasets} 

\begin{table}
\caption{\small Statistics of evaluation datasets}\label{table:datasets}
\small
  \setlength{\tabcolsep}{2pt}
\centering
\begin{tabular}{|l|c|c|c|c|c|} \hline
\textbf{Dataset}&\textbf{\#entities}&\textbf{\#relations}&\textbf{\#train}&\textbf{\#valid}&\textbf{\#test}\\ \hline
\textbf{FB15k}&14,951&1,345&483,142&50,000&59,071\\ \hline
\textbf{FB15k-237}&14,541&237&272,115&17,535&20,046\\ \hline
\textbf{WN18}&40,943&18&141,442&5,000&5,000\\ \hline
\textbf{WN18RR}&40,943&11&86,835&3,034&3,134\\ \hline
\textbf{YAGO3-10}&123,182&37&1,079,040&5,000&5,000\\ \hline
\textbf{YAGO3-10-DR}&122,837&36&732,556&3,390&3,359\\ \hline
\end{tabular}
\end{table}

{\flushleft \textbf{FB15k}:}\hspace{2mm} Most embedding models have been evaluated on FB15k, a subset of Freebase generated by Bordes et al.~\cite{bordes2013translating}. 
\gotoTR{
Freebase~\cite{freebase} is one of the largest public domain knowledge bases. Its final 2015 version records 1.9 billion triples of common facts. This knowledge base, available in various data formats including RDF, can be modeled as a graph in which the entities and relations correspond to nodes and edges, respectively.
}
FB15k contains only those Freebase entities that were also available in Wikipedia based on the \emph{wiki-links} database~\footnote{\small{\url{https://code.google.com/archive/p/wiki-links/}}} and have at least 100 appearances in Freebase. The relations included into FB15k must also have at least 100 instances. 14,951 entities and 1,345 relations satisfy these criteria, which account for 592,213 triples included into FB15k. These triples were randomly split into training, validation and test sets. Table~\ref{table:datasets} shows the statistics of this and other datasets. \vspace{-1mm}

{\flushleft \textbf{WN18}:}\hspace{2mm}
Many embedding models have also been evaluated using WN18~\cite{bordes2013translating}, a knowledge graph extracted from the English lexical database WordNet~\cite{wordnet} which defines conceptual-semantic and lexical relations between word forms or between synsets---sets of synonyms. 
An example triple is \triple{presentation}{derivationally\_related\_form}{present}. \vspace{-1mm}

{\flushleft \textbf{YAGO3-10}:}\hspace{2mm} 
Some embedding models were evaluated using YAGO3-10~\cite{ConvE}, a subset of YAGO3~\cite{yago3}---the multilingual extension of YAGO~\cite{yago} which is derived from Wikipedia and WordNet. YAGO3-10 contains entities that are involved in at least 10 relations in YAGO3. 
\vspace{-1mm}

\vspace{-2mm}
\subsection{Evaluation Methods and Measures}\label{sec:eval-measure}

Embedding models have been evaluated using several highly-related knowledge graph completion tasks such as triple classification~\cite{TransH,socher2013reasoning}, link prediction~\cite{PTransE}, relation extraction~\cite{weston,TransR}, and relation prediction~\cite{proje}. The \emph{link prediction} task as described in~\cite{bordes2013translating} is particularly widely used for evaluating different embedding methods. Its goal is to predict the missing \entity{h} or \entity{t} in a triple \triple{h}{r}{t}. For each test triple \triple{h}{r}{t}, the head entity \entity{h} is replaced with every other entity \(\entity{h'} \in \mathcal{E}\) in the dataset, to form \emph{corrupted} triples. 
The original test triple and its corresponding corrupted triples are ranked by their scores according to the score functions (Section~\ref{sec:embedding}) and the rank of the original test triple is denoted \emph{rank}$_\entity{h}$.  The same procedure is used to calculate \emph{rank}$_\entity{t}$ for the tail entity \entity{t}.  A method with the ideal ranking function should rank the test triple at top.

The accuracy of different embedding models is measured using \hitone, \hitten, Mean Rank (\mr), and Mean Reciprocal Rank (\mrr), as in~\cite{bordes2013translating}. 
\texttt{\small Hits@k$^\uparrow$} is the percentage of top $k$ results that are correct. \mr\ is the mean of the test triples' ranks, defined as $MR=\frac{1}{2\ |T|}\sum_{\triple{h}{r}{t} \in T}(\emph{rank}_\entity{h} + \emph{rank}_\entity{t})$, 
in which \(|T|\) is the size of the test set. \mrr\ is the average inverse of harmonic mean of the test triples' ranks, defined as $MRR=\frac{1}{2\ |T|}\sum_{\triple{h}{r}{t} \in T}(\frac{1}{\emph{rank}_\entity{h}} + \frac{1}{\emph{rank}_\entity{t}})$. 

Besides these raw metrics, we also used their corresponding \emph{filtered} metrics~\cite{bordes2013translating}, denoted \fhitone, \fhitten, \fmr, and \fmrr, respectively. In calculating these measures, corrupted triples that are already in training, test or validation sets do not participate in ranking. In this way, a model is not penalized for ranking other correct triples higher than a test triple. For example, consider the task of predicting tail entity. Suppose the test triple is \triple{Tim Burton}{film}{Edward Scissorhands} and the training, test, or validation set also contains another triple \triple{Tim Burton}{film}{Alice in Wonderland}. If a model ranks \entity{Alice in Wonderland} higher than \entity{Edward Scissorhands}, the filtered metrics will remove this film from the ranked list so that the model would not be penalized for ranking \triple{Tim Burton}{film}{Edward Scissorhands} lower than \triple{Tim Burton}{film}{Alice in Wonderland}, both correct triples. 

We note that, by definition, higher \hitone\ (\fhitone), \hitten\ (\fhitten) and \mrr\ (\fmrr{}), and lower \mr\ (\fmr{}) indicate better accuracy. 

%% file: sec-issues.tex
\section{Inadequacy of benchmarks and Evaluation Measures}\label{sec:issues}

In this section we investigate the existence and impact of a few types of relations in FB15k, WN18 and YAGO3-10, including reverse (and symmetric) relations, redundant relations, and Cartesian product relations. The outcome suggests that triples in these relations led to a substantial over-estimation of the accuracy of embedding models.

\input{sec-create.tex}

\subsection{Data Redundancy}\label{sec:redundancy}

\subsubsection{Data Leakage Due to Reverse Triples}\label{sec:reverserelation}

(1) \textbf{FB15k}: In~\cite{toutanova2015observed}, Toutanova and Chen noted that the widely-used benchmark dataset FB15k contains many reverse triples, i.e., it includes many pairs of \triple{h}{r}{t} and \triple{t}{r$^{-1}$}{h} where \rel{r}$^{-1}$ is the reverse of \rel{r}. Freebase actually denotes reverse relations explicitly using a special relation \rel{reverse\_property}.\footnote{The relation's full name is \rel{/type/property/reverse\_property}. In Freebase, the full name of a relation follows the template of /domain/entity type/relation. 
For instance, \rel{/tv/tv\_genre/programs} represents a relation named \emph{programs} belonging to domain \emph{tv}. The subject of any instance triple of this relation belongs to type \emph{tv\_genre}. For simplicity of presentation, by default we omit the prefixes. In various places we retain the entity type to avoid confusions due to identical relation names.} For instance, the triple \triple{film/directed\_by}{reverse\_property}{director/film} in Freebase denotes that \rel{film/directed\_by} and \rel{director/film} are reverse relations.\footnote{Note that relations \rel{film/directed\_by} and \rel{director/film} are also special entities in \triple{film/directed\_by}{reverse\_property}{director/film}.} 
Therefore, \triple{A Room With A View}{film/directed\_by}{James Ivory} and \triple{James Ivory}{director/film}{A Room With A View} form a pair of reverse triples, as shown in Figure~\ref{fig:mid node}. The reverse relation of a concatenated relation $\rel{r}_1.\rel{r}_2$ is the concatenation of the corresponding reverse relations, i.e.,
$\rel{r}_2^{-1}.\rel{r}_1^{-1}$. For example, in Figure~\ref{fig:mid node} \rel{award\_nominated\_work/award\_nominations\ .\ award\_nomination/award} (blue edges) and \rel{award\_category/nominees\ .\ award\_nomination/nominated\_for} (red edges) are two concatenated relations which are reverse of each other.

Using the reverse relation information from the May 2013 snapshot of Freebase, out of the 483,142 triples in the training set of FB15k, 338,340 triples form 169,170 reverse pairs. Furthermore, for 41,529 out of the 59,071 triples (i.e., about 70.3\%) in the test set of FB15k, their reverse triples exist in the training set. These data characteristics suggest that embedding models would have been biased toward learning reverse relations for link prediction. More specifically, the task can largely become inferring whether two relations \rel{r}$_1$ and \rel{r$_2$} form a reverse pair. Given the abundant reverse triples in the dataset, this goal could potentially be achieved without using a machine learning approach based on complex embeddings of entities and relations.  Instead, one may aim at deriving simple rules of the form \triple{h}{r$_1$}{t} $\Rightarrow$ \triple{t}{r$_2$}{h} using statistics about the triples in the dataset. In fact, Dettmers et al.~\cite{ConvE} generated such a simple model which attained a 68.9\% accuracy by the measure \fhitone. We generated a similar model by finding the relations that have more then 80\% intersections. It attained an \fhitone\ of 71.6\%. This is even slightly better than the 70.3\% accuracy one may achieve using an oracle based on the reverse relations denoted in the May 2013 Freebase snapshot. The \fhitone\ of the best performing embedding model on FB15k is 73.8\% (more details in Table~\ref{table:Hits@1} of Section~\ref{sec:exp}). 

(2) \textbf{WN18}: WN18 also suffers from data leakage, as 14 out of its 18 relations form 7 pairs of reverse relations, e.g., \triple{europe}{has\_part}{republic\_of\_estonia} and \triple{republic\_of\_estonia}{part\_of}{europe} are two reverse triples in reverse relations \rel{has\_part} and \rel{part\_of}. There are also 3 self-reciprocal (i.e., symmetric) relations: \rel{verb\_group}, \rel{similar\_to}, \rel{derivationally\_related\_form}. 4,658 out of the 5,000 test triples have their reverse triples available in the training set. The training set itself contains 130,791 (about 92.5\%) triples that are reverse of each other. On WN18, we can achieve an \fhitone of 96.4\% by the aforementioned simple rule-based model (finding the relations that have more then 80\% intersections) which is better than the results obtained by the embedding models (Table~\ref{table:Hits@1} of Section~\ref{sec:exp}).

In training a knowledge graph completion model using FB15k and WN18, we fall into a form of overfitting in that the learned models are optimized for the reverse triples which cannot be generalized to realistic settings. More precisely, this is a case of excessive data leakage---the model is trained using features that otherwise would not be available when the model needs to be applied for real prediction.

\begin{figure}
    \centering
    \includegraphics[trim={2.2cm 19cm 7.2cm 3cm},clip,scale=0.5,width=0.40\textwidth]{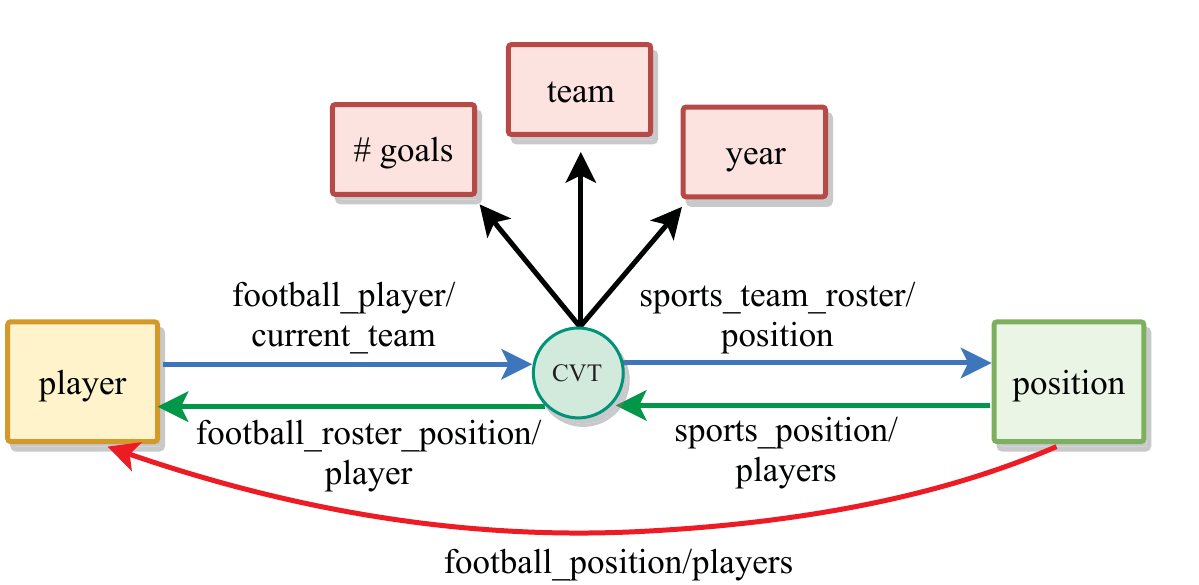}
    \caption{\small Duplicate relations}
    \label{fig:duplicate}
\end{figure}

\subsubsection{Other Redundant Triples}\label{sec:duplicate}

(1) \textbf{FB15k}: 
In addition to reverse relations, there are other types of semantically redundant relations in FB15k. While it is infeasible to manually verify such semantic redundancy, we used a simple method to automatically detect it. Given two relations $\rel{r}_1$ and $\rel{r}_2$, we calculate how much their subject-object pairs overlap. Suppose $|\rel{r}|$ is the number of instance triples in relation \rel{r} and $T_\rel{r}$ denotes the set of subject-object pairs in \rel{r}, i.e., $T_\rel{r}=\{(\entity{h}, \entity{t})\ |\ \rel{r}(\entity{h},\entity{t}) \in \mathcal{G} \}$. We say $\rel{r}_1$ and $\rel{r}_2$ are \emph{near-duplicate relations}, simplified as \emph{duplicate relations}, if they satisfy the following condition: $\frac{|T_{\rel{r}_1} \cap T_{\rel{r}_2}|}{|\rel{r}_1|} > \theta_1$ and $\frac{|T_{\rel{r}_1} \cap T_{\rel{r}_2}|}{|\rel{r}_2|} > \theta_2$. 
Moreover, $T_\rel{r}^{-1}$ denotes the reverse entity pairs of $T_\rel{r}$, i.e., $T_\rel{r}^{-1} = \{(\entity{t}, \entity{h})\ |\ (\entity{h}, \entity{t})$ $\in$ $T_\rel{r} \}$. We say $\rel{r}_1$ and $\rel{r}_2$ are \emph{reverse duplicate relations} if they satisfy the following condition: $\frac{|T_{\rel{r}_1} \cap T_{\rel{r}_2}^{-1}|}{|\rel{r}_1|} > \theta_1$ and $\frac{|T_{\rel{r}_1} \cap T_{\rel{r}_2}^{-1}|}{|\rel{r}_2|} > \theta_2$. 
We have set $\theta_1$ and $\theta_2$ to 0.8 on FB15k.

For example \rel{football\_position/players} ($\rel{r}_1$) and \rel{sports\_position\allowbreak /players.football\_roster\_position/player} ($\rel{r}_2$) are duplicate based on this definition, since $\frac{|T_{\rel{r}_1} \cap T_{\rel{r}_2}|}{|\rel{r}_1|} = 0.87$ and $\frac{|T_{\rel{r}_1} \cap T_{\rel{r}_2}|}{|\rel{r}_2|} = 0.97$. These two relations are displayed in Figure~\ref{fig:duplicate} using red and green edges, respectively. The first relation records each football player's position considering their overall career. For the second relation, each instance triple is a concatenation of two edges connected through a mediator node, representing a multiary relationship about the position a player plays for a team as shown in Figure~\ref{fig:duplicate}. Since most players play at the same position throughout their careers, these two relations are redundant. Another example of similar nature is that $\rel{r}_1$ and \rel{football\_player/current\_team~.~sports\_team\_roster/position} ($\rel{r}_3$) are reverse duplicate relations, because 
$\frac{|T_{\rel{r}_1} \cap T_{\rel{r}_3}^{-1}|}{|\rel{r}_1|} = 0.88$ and $\frac{|T_{\rel{r}_1} \cap T_{\rel{r}_3}^{-1}|}{|\rel{r}_3|} = 0.97$. In Figure~\ref{fig:duplicate} they are highlighted in red and blue, respectively. 

\begin{figure}
    \centering
    \includegraphics[trim={3cm 12cm 6cm 2cm},clip,width=0.23\textwidth]{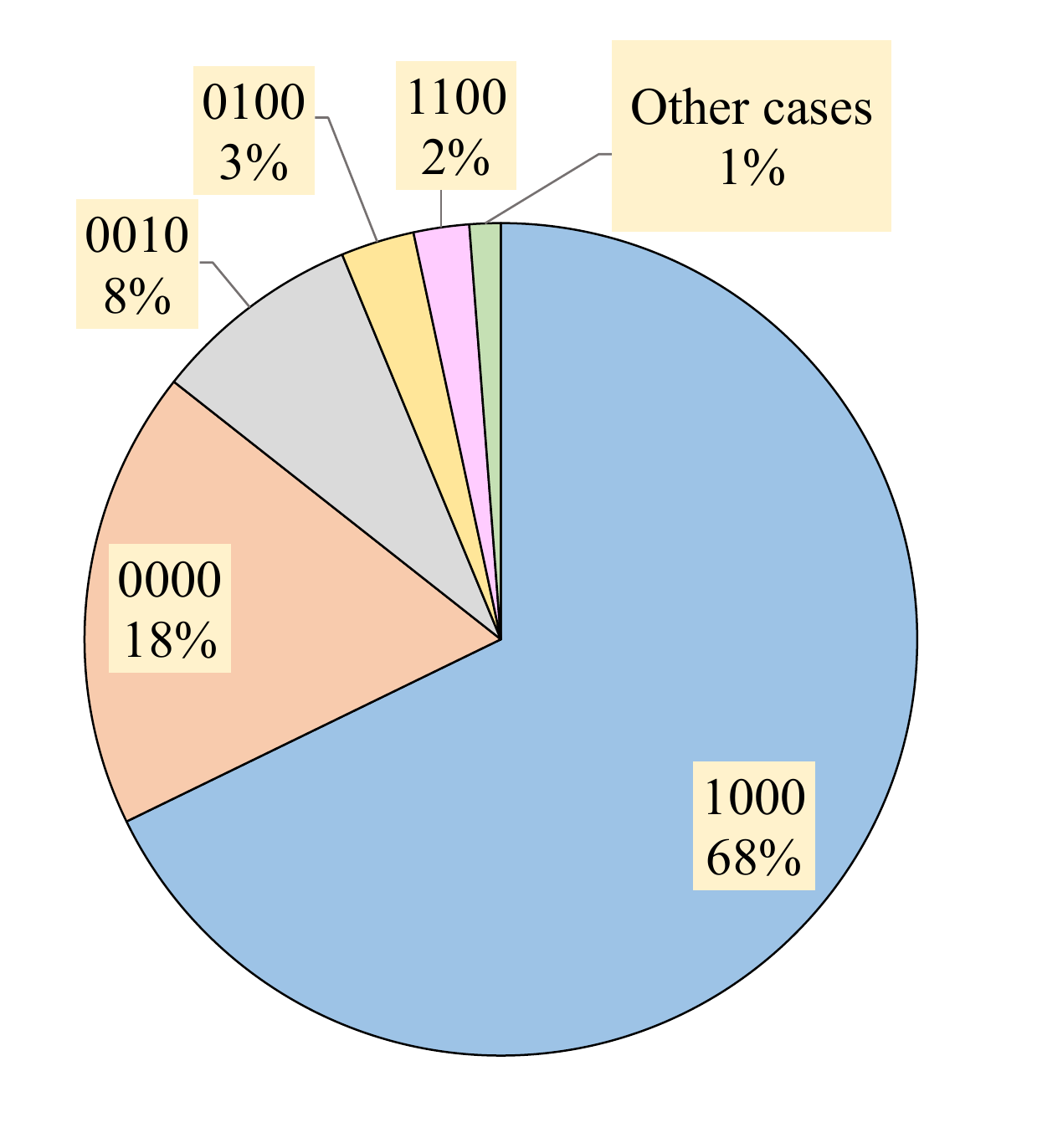}
    \vspace{-2mm}
    \caption{\small Redundancy in the test set of FB15k}
    \label{fig:test_redundancies}
\end{figure}

For each test triple in FB15k, we use the methods explained to determine whether it has 1) reverse triples, 2) duplicate or reverse duplicate triples in the training set, and whether it has 3) reverse triples, 4) duplicate or reverse duplicate triples in the test set itself. A triple may have redundant triples in any of these four categories. We use bitmap encoding to represent different cases of redundancy. For example, 1100 is for a triple that has both reverse triples and  (reverse) duplicate triples in the training set. Hence, there are 16 possible different combinatorial cases. Considering the test triples in FB15k, not all 16 cases exist. Instead, 12 different cases exist. Figure~\ref{fig:test_redundancies} shows the percentages of triples in different cases. The 7 cases smaller than 1\% are combined in one slice. The largest three slices are 1000 (triples with only reverse triples in the training set), 0000 (triples without any redundant triples), and 0010 (triples with only reverse triples in the test set). In total, 41,529, 1,847 and 2,701 test triples have reverse, reverse duplicate and duplicate triples in the training set, and 4,992, 249, and 328 test triples have these categories of redundant triples in the test set itself. The data redundancy causes overestimation of embedding models' accuracy. For instance, the \fmr, \fhitten, \fhitone, and \fmrr\ of ConvE are 33.5, 88.8, 64.3, and 0.734 on such relations, whereas its performance using these measures on relations without any redundancy is only 149, 61.2, 37.3, and 0.454, respectively.

(2) \textbf{YAGO3-10}: With 1,079,040 training triples, this dataset is larger than FB15k and WN18.  However, among its 37 relations, the two most populated relations $\rel{isAffiliatedTo}\ (r_1)$ and $\rel{playsFor}\ (r_2)$ account for 35\% and 30\% of the training triples, respectively. Although $r_1$ semantically subsumes $r_2$ in the real world, they appear as near-duplicate relations in this particular dataset, as $\frac{|T_{\rel{r}_1} \cap T_{\rel{r}_2}|}{|\rel{r}_1|} = 0.75$ and $\frac{|T_{\rel{r}_1} \cap T_{\rel{r}_2}|}{|\rel{r}_2|} = 0.87$.  In the training set, 557,696 triples find their duplicates in itself. 2,561 out of the 5,000 test triples have their duplicate triples available in the training set. The various models achieved much stronger results on $r_1$ and $r_2$ than other relations. For example, the \fmr, \fhitten, \fhitone, and \fmrr\ of RotatE on these two relations are 225.84, 81.04, 50.43, and 0.612, in comparison with 4,540.65, 43.76, 23.38, and 0.304 on other relations. Furthermore, YAGO3-10 also has 3 semantically symmetric relations: \rel{hasNeighbor}, \rel{isConnectedTo}, and \rel{isMarriedTo}. In its test set, 118 triples belonging to these relations have their reverse triples available in the training set.  The \fhitone of the simple rule-based model mentioned in Section~\ref{sec:intro} is 51.6\% on YAGO3-10, which outperforms embedding models, as can be seen in Table~\ref{table:results yago}.

\begin{table*}
\caption{\small The strong \fmrr results on a few Cartesian product relations in FB15k-237}\label{table:cart}
\small
\centering
\setlength{\tabcolsep}{2pt}
\begin{tabular}{|c|c|c|c|c|c|c|} \hline
\textbf{relation} & \textbf{\# of triples} & \textbf{TransE} & \textbf{DistMult} & \textbf{ComplEx} &  \textbf{ConvE} & \textbf{RotatE}\\ \hline

\makecell{\rel{olympic\_games/medals\_awarded . olympic\_medal\_honor/medal}}  & 16& 1& 1& 1&  1 &1\\ \hline

\makecell{\rel{food/nutrients . nutrition\_fact/nutrient}}& 105 & 0.83&0.73&0.72&0.82 &0.79\\ \hline 

\makecell{\rel{travel\_destination/climate . travel\_destination\_monthly\_climate/month}}& 60 & 0.98&0.77&0.95&0.98&0.98\\ \hline

\end{tabular}
\end{table*}

\begin{table*}
\caption{\small Link prediction using Cartesian product property}\label{table:cartesian}
\scriptsize
    \centering
    \setlength{\tabcolsep}{2pt}
    \begin{tabular}{|l|c|c|c|c|c|c|c|c||c|c|c|c|c|c|c|c|c|c|c|c|}\hline
    &\multicolumn{8}{c||}{\textbf{\small TransE results}}&\multicolumn{12}{c|}{\textbf{\small Prediction using Cartesian product property} }\\\hline
    &\multicolumn{8}{c||}{ \textbf{\small FB15k as ground truth}} & \multicolumn{8}{c|}{ \textbf{\small FB15k as ground truth}}& \multicolumn{4}{c|}{\textbf{\small Freebase as ground truth}}\\\hline
          & \textbf{\mr} & \textbf{\hten} & \textbf{\hone} & \textbf{\mrr}& \textbf{\fmr} & \textbf{\fhten} & \textbf{\fhone} & \textbf{\fmrr}& \textbf{\mr} & \textbf{\hten} & \textbf{\hone} & \textbf{\mrr}& \textbf{\fmr} & \textbf{\fhten} & \textbf{\fhone} & \textbf{\fmrr}& \textbf{\fmr} & \textbf{\fhten} & \textbf{\fhone} & \textbf{\fmrr}\\\hline
           \textbf{r1} & 19.97 & 43.33 & 3.33 & 0.14 & 1.39 & 99.17 & 83.33 & 0.9& 
            16.52   & 52  & 6  & 0.18 & 1.46   & 100 & 70  & 0.83 & 1.40   & 100 & 71  & 0.84\\\hline
            
            \textbf{r2}& 5 & 100 & 0  & 0.24 & 2.5 & 100 & 0 & 0.42&
              3.6       & 100 & 35  & 0.55 & 1.6     & 100 & 70  & 0.80 & 1.45      & 100 & 75 & 0.84\\\hline
                                     
            \textbf{r3}& 6.5 & 100 & 0  & 0.22 & 2.5 & 100 & 0 & 0.42&
              3.6     & 100 & 40  & 0.57 & 1.45      & 100 & 70 & 0.82    & 1.4      & 100 & 70 & 0.83\\\hline
            
            \textbf{r4}&1784.25 & 25 & 0 & 0.09 & 1777 & 25 & 0 & 0.1  &
            995.25 & 74  & 25 & 0.45 & 990.17   & 88  & 88  & 0.88 & 990.09   & 88  & 88  & 0.88\\\hline
            
            \textbf{r5}&12.74 & 58.82 & 11.76 & 0.31 & 1 & 100 & 100 & 1&
            364.03  & 60  & 17 & 0.35 & 352.03    & 94  & 94  & 0.94 & 352.03    & 94  & 94  & 0.94 \\\hline
            
            \textbf{r6}&10.72 & 62.5 & 9.38 & 0.27 & 1.03 & 100 & 96.88 & 0.98 &
            9.62    & 64  & 12 & 0.31 & 1      & 100 & 100 & 1    & 1      & 100 & 100 & 1 \\\hline
            
            \textbf{r7}&7.75 & 50 & 25 & 0.35 & 2.75 & 100 & 25 & 0.51 &
            1872.38 & 60  & 10  & 0.3 & 1865.72 & 75  & 25  & 0.43 & 1864.83 & 75  & 53  & 0.6\\\hline
            
            \textbf{r8}&10.75 & 62.5 & 0 & 0.19 & 4.63 & 75 & 25 & 0.44 &
            6.63    & 74  & 14  & 0.37 & 2.69   & 100 & 60  & 0.7 & 2.09    & 100 & 64  & 0.74\\\hline
            
            \textbf{r9}&15.59 & 53.13 & 18.75 & 0.34 & 1 & 100 & 100 & 1 &
            11.57   & 65  & 19 & 0.36 & 1      & 100 & 100 & 1    & 1      & 100 & 100 & 1\\\hline
    \end{tabular}
   
\end{table*}
\begin{table}
\caption{\small Cartesian product relations used in Table~\ref{table:cartesian}}
    \centering
     \setlength{\tabcolsep}{2pt}
    \begin{tabular}{|c|c|}\hline
         \textbf{r1}& \scriptsizerel{travel\_destination/climate~.~travel\_destination\_monthly\_climate/month} \\\hline
         \textbf{r2}& \scriptsizerel{computer\_videogame/gameplay\_modes} \\\hline
         \textbf{r3}& \scriptsizerel{gameplay\_mode/games\_with\_this\_mode}\\\hline
         \textbf{r4}& \scriptsizerel{educational\_institution/sexes\_accepted . gender\_enrollment/sex}\\\hline
         \textbf{r5}& \scriptsizerel{olympic\_medal/medal\_winners . olympic\_medal\_honor/olympics}\\\hline
         \textbf{r6}& \makecell[c]{\scriptsizerel{x2010fifaworldcupsouthafrica/world\_cup\_squad/current\_world\_cup\_squad . }
         \\\scriptsizerel{x2010fifaworldcupsouthafrica/current\_world\_cup\_squad/position}}\\\hline
         \textbf{r7}& \scriptsizerel{dietary\_restriction/compatible\_ingredients}\\\hline
         \textbf{r8}& \scriptsizerel{ingredient/compatible\_with\_dietary\_restrictions}\\\hline
         \textbf{r9}& \scriptsizerel{olympic\_games/medals\_awarded . olympic\_medal\_honor/medal}\\\hline
    \end{tabular}
    \label{table:cartesian-rel}
\end{table}

\subsection{Cartesian Product Relations}\label{sec:cartesian}

We also discovered another issue with FB15k which makes existing performance measures of embedding models unrealistic. 
This problem manifests in what we call \emph{Cartesian product relations}. Given such a relation, the subject-object pairs from all instance triples of the relation form a Cartesian product. In other words, there are a set of subjects and a set of objects, and the relation exists from every subject in the first set to every object in the second set. One example Cartesian product relation is \rel{climate}, since \triple{a}{climate}{b} is a valid triple for every possible city \entity{a} and month \entity{b}. Another example is \rel{position}, since every team in a certain professional sports league has the same set of positions. The link prediction problem for such relations thus becomes predicting whether a city has a climate in, say, January, or whether an NFL team has the quarter-back  position. Such a prediction task is not very meaningful. 

A few notes can be made about Cartesian product relations. (1) Similar to reverse relations and other forms of data redundancy, the existence of these relations unrealistically inflates a model's link prediction accuracy. When a substantial subset of the aforementioned subject-object Cartesian product is available in the training set, it is relatively easy for a model to attain strong accuracy. However, it is problematic to mix such straightforward test cases with more realistic, challenging cases. At least, the performance of a model should be separately evaluated on Cartesian product relations and non-Cartesian product relations. 

(2) Albeit not always meaningful, one may still perform link prediction on Cartesian product relations. However, a simpler approach can be more effective than learning complex embedding models. For instance, by examining all instance triples of a relation \rel{r}, a method can identify the sets of all subjects $S_\rel{r} =\{ \entity{h}\ |\ \exists \rel{r}(\entity{h},\entity{t}) \in \mathcal{G}\}$ and objects $O_\rel{r} =\{ \entity{t}\ |\ \exists \rel{r}(\entity{h},\entity{t}) \in \mathcal{G}\}$ in the instance triples. By observing a large percentage of possible subject-object pairs existing in the relation, the method can derive the relation might be a Cartesian product relation. More specifically, if \(|\rel{r}|\ /\ (|S_\rel{r}| \times |O_\rel{r}|) \) is greater than a pre-determined threshold (0.8 in our study), we consider \rel{r} a Cartesian product relation. 
There are a total of 31,771 relations (375,387,927 instance triples) in the aforementioned May 2013 Freebase snapshot, of which 3,568 relations have only one instance triple each. Among the 28,203 remaining relations (375,384,359 instance triples), we detected 2,951 Cartesian product relations (2,605,338 instance triples) using this method. We also identified 142 Cartesian product relations in FB15k, with 13,038 triples. 
Although there are not as many Cartesian product relations as reverse relations, we discovered that among the relations on which embedding models attained the highest accuracy there are Cartesian product relations. Table~\ref{table:cart} shows such results on the relations using \fmrr\ on FB15k after reverse triples are removed.~\footnote{This dataset is called FB15k-237 and will be further discussed in Section~\ref{sec:exp}. We want to inspect the impact of Cartesian product relations after removing reverse relations, because the latter also causes over-estimation of embedding models' accuracy and they dominate Cartesian product relations in terms of number of triples.} These are the Cartesian product relations among the top-12 relations ranked by \fmrr\ of ConvE on FB15k-237. 

Once a relation \rel{r} is detected as a Cartesian product relation, for link prediction, we can predict triple \triple{h}{r}{t} to be valid, given any  $\entity{h} \in S_\rel{r}$ and $\entity{t} \in O_\rel{r}$.  We can further extend this approach. If an entity type system exists (which is the case with Freebase), we can identify the common type of all entities in $S_\rel{r}$ and $O_\rel{r}$, respectively, and then predict \triple{h}{r}{t} valid for all $\entity{h}$ and $\entity{t}$ belonging to the corresponding types.  

(3) The existence of Cartesian product relations in FB15k is quite artificial. In fact, many such relations are due to mediator nodes in Freebase and simplification in FB15k for removing such nodes (see Section~\ref{sec:snapshot}). The majority of relationships in Freebase are multiary relationships connected through mediator nodes. For instance, a mediator node is connected to \entity{Tokyo} with an edge labeled \rel{climate} and to \entity{January} with an edge labeled \rel{month}. It is further connected to \entity{34} with an edge labeled \rel{average\_min\_temp\_c}, indicating that the average low temperature in Tokyo is 34 degrees Fahrenheit in January. In fact, it is also connected to other nodes for capturing the maximal temperature, rain fall, and so on. 
The more realistic and useful prediction task is to predict the average temperature, rather than whether a city has a temperature. Even though most real-world relationships are multiary, the studies on link prediction have often simplified it as multiple binary relationships (which is lossful as the multiple binary relationships cannot be used to restore the identical original relationship). That is how FB15k entails the less meaningful prediction tasks. In fact, out of the 4,683 Cartesian product relations mentioned in (2), 3,506 are concatenated relations. In the specific example above, the concatenated edge is between \entity{Tokyo} and \entity{January}, connecting the original \rel{climate} and \rel{month} edges. 

\bgroup
\begin{table*}
\caption{\small Link prediction results on FB15k and FB15k-237}\label{table:results}
\scriptsize
  \setlength{\tabcolsep}{2pt}
  \begin{tabular}{|P{68pt}|P{24pt}|P{38pt}|P{25pt}|P{25pt}|P{42pt}|P{29pt}||P{24pt}|P{38pt}|P{25pt}|P{25pt}|P{42pt}|P{29pt}|}
  \hline
    
\multicolumn{7}{|c||}{\textbf{\small FB15k}}&\multicolumn{6}{c|}{\textbf{\small FB15k-237}}  \\\hline
&\multicolumn{3}{c|}{\textbf{\small Raw measures}}&\multicolumn{3}{c||}{\textbf{\small Filtered measures}}&\multicolumn{3}{c|}{\textbf{\small Raw measures}}&\multicolumn{3}{c|}{\textbf{\small Filtered measures}}\\\hline
 \textbf{\small Model} & \textbf{\mr} & \textbf{\hitten} & \textbf{\mrr} & \textbf\fmr& \textbf\fhitten&\textbf\fmrr &  \textbf\mr & \textbf\hitten &\textbf\mrr& \textbf\fmr & \textbf\fhitten &\textbf\fmrr    \\ \hline

\textbf{TransE \cite{bordes2013translating}}   & 
\makecell[c]{243.0 \\ \color{blue}199.9} & 
\makecell[c]{34.9 \\ \color{blue}44.3} & 
\makecell[c]{-- \\ \color{blue}0.227} & 
\makecell[c]{125.0\\ \color{blue}68.8} &
\makecell[c]{47.1 \\\color{blue}62.4} & 
\makecell[c]{--\\ \color{blue}0.391} &

\makecell[c]{--\\ \color{blue}363.3} & 
\makecell[c]{--\\ \color{blue}32.2} &  
\makecell[c]{--\\\color{blue}0.169} & 
\makecell[c]{--\\ \color{blue}223.4} &
\makecell[c]{--\\\color{blue}47.5} &
\makecell[c]{--\\\color{blue}0.288}\\ \hline

\textbf{TransH \cite{TransH}} & 
\makecell[c]{211.0  \\\color{blue}234.7} & 
\makecell[c]{42.5\\\color{blue}45.5} &
\makecell[c]{--\\\color{blue}0.177 }&
\makecell[c]{84.0\\ \color{blue}84.0} & 
\makecell[c]{58.5\\\color{blue}69.0} &
\makecell[c]{--\\\color{blue}0.346}&

\makecell[c]{--\\\color{blue}398.8} & 
\makecell[c]{--\\\color{blue}30.9}& 
\makecell[c]{--\\\color{blue}0.157}& 
\makecell[c]{--\\\color{blue}251.1}&
\makecell[c]{--\\\color{blue}49.0}&  
\makecell[c]{--\\\color{blue}0.290}\\ \hline

\textbf{TransR \cite{TransR}}  &
\makecell[c]{226.0 \\\color{blue}231.9} &
\makecell[c]{43.8\\ \color{blue}48.8}&
\makecell[c]{--\\\color{blue}0.236 }&
\makecell[c]{78.0\\  \color{blue}78.2} & 
\makecell[c]{65.5\\ \color{blue}72.9}&
\makecell[c]{--\\  \color{blue}0.471}&

\makecell[c]{--\\ \color{blue}391.3} &
\makecell[c]{--\\\color{blue} 31.4} & 
\makecell[c]{--\\ \color{blue}0.164} &
\makecell[c]{-- \\ \color{blue}240.2}  &
\makecell[c]{--\\ \color{blue}51.0} &
\makecell[c]{--\\ \color{blue}0.314} \\ \hline

\textbf{TransD \cite{TransD}}  &
\makecell[c]{211.0\\ \color{blue}234.4} & 
\makecell[c]{49.4 \\ \color{blue}47.4} & 
\makecell[c]{-- \\ \color{blue}0.179 }& 
\makecell[c]{67.0\\  \color{blue}85.4}&
\makecell[c]{74.2\\ \color{blue}70.9} &
\makecell[c]{--\\ \color{blue}0.352}&

\makecell[c]{--\\ \color{blue}391.6 }&
\makecell[c]{--\\ \color{blue}30.6} &
\makecell[c]{--\\ \color{blue}0.154}  &
\makecell[c]{--\\ \color{blue}244.1} &
\makecell[c]{--\\ \color{blue}48.5} &
\makecell[c]{--\\ \color{blue}0.284 }\\ \hline




 \textbf{DistMult \cite{DistMult} } &
 \makecell[c]{-- \\ \color{blue}313.0\\ \color{green}264.3 } & 
 \makecell[c]{--\\ \color{blue}45.1\\ \color{green}50.3 } &
 \makecell[c]{-- \\ \color{blue}0.206\\ \color{green}0.240}& 
 \makecell[c]{-- \\ \color{blue}159.6\\ \color{green}106.3 }  &
 \makecell[c]{57.7\\ \color{blue}71.4\\  \color{green}82.8 }  & 
 \makecell[c]{0.35 \\  \color{blue}0.423\\  \color{green}0.651} & 
 
 \makecell[c]{-- \\ \color{blue}566.3\\ \color{green}--} & 
 \makecell[c]{-- \\ \color{blue} 30.3\\ \color{green}-- }& 
 \makecell[c]{-- \\ \color{blue}0.151\\\color{green}-- }&
 \makecell[c]{--\\ \color{blue}418.5\\ \color{green}--}&
 \makecell[c]{-- \\ \color{blue}41.8\\\color{green}--} &
 \makecell[c]{-- \\ \color{blue}0.238\\\color{green}--}\\ \hline

\textbf{ComplEx \cite{complex}}  &
\makecell[c]{-- \\ \color{blue}350.3\\ \color{green}250.6} &
\makecell[c]{--\\ \color{blue}43.8\\ \color{green}49.2 } &
\makecell[c]{0.242 \\ \color{blue}0.205\\ \color{green}0.233} & 
\makecell[c]{--\\ \color{blue}192.3\\ \color{green}90.0} &
\makecell[c]{84.0\\ \color{blue}72.7\\ \color{green}83.2}&
\makecell[c]{0.692\\  \color{blue}0.516\\ \color{green}0.685}&

\makecell[c]{--\\ \color{blue}656.4\\\color{green}-- }&
\makecell[c]{--\\ \color{blue}29.9\\\color{green}--}&
\makecell[c]{-- \\ \color{blue}0.158\\\color{green}-- } &
\makecell[c]{--\\ \color{blue}508.5\\ \color{green}--} & 
\makecell[c]{--\\ \color{blue}42.3\\\color{green}--} & 
\makecell[c]{--\\ \color{blue}0.249\\\color{green}--}\\ \hline 

\textbf{ConvE \cite{ConvE}}  &
\makecell[c]{--\\ \color{orange}189.5}  &
\makecell[c]{--\\\color{orange}51.7 }& 
\makecell[c]{--\\\color{orange}0.268} &
\makecell[c]{64.0\\\color{orange}46.5} &
\makecell[c]{87.3\\\color{orange}85.6} &
\makecell[c]{0.745\\\color{orange}0.698}&

\makecell[c]{--\\\color{orange}481.7}   & 
\makecell[c]{-- \\\color{orange}28.6}  &
\makecell[c]{--\\\color{orange}0.154} &
\makecell[c]{246.0 \\ \color{orange}271.5}&
\makecell[c]{49.1 \\ \color{orange}48.1}  & 
\makecell[c]{0.316 \\ \color{orange}0.305}\\ \hline

\textbf{RotatE \cite{RotatE}}  &
\makecell[c]{--\\ \color{cyan}190.4}  &
\makecell[c]{--\\\color{cyan}50.6}& 
\makecell[c]{--\\\color{cyan}0.256} &
\makecell[c]{40\\\color{cyan}41.1} &
\makecell[c]{88.4\\\color{cyan}88.1} &
\makecell[c]{0.797\\\color{cyan}0.791}&

\makecell[c]{--\\\color{cyan}333.4}   & 
\makecell[c]{-- \\\color{cyan}31.7}  &
\makecell[c]{--\\\color{cyan}0.169} &
\makecell[c]{177 \\ \color{cyan}179.1}&
\makecell[c]{53.3 \\ \color{cyan}53.2}  & 
\makecell[c]{0.338 \\ \color{cyan}0.337}\\ \hline

\textbf{TuckER \cite{TuckER}}  &
\makecell[c]{--\\ \color{pink}186.4}  &
\makecell[c]{--\\\color{pink} 51.3}& 
\makecell[c]{--\\\color{pink}0.260} &
\makecell[c]{--\\\color{pink}39.0} &
\makecell[c]{89.2\\\color{pink}89.1} &
\makecell[c]{0.795\\\color{pink}0.790}&

\makecell[c]{--\\\color{pink}343.5}   & 
\makecell[c]{-- \\\color{pink}35.4}  &
\makecell[c]{--\\\color{pink}0.197} &
\makecell[c]{-- \\ \color{pink}164.8}&
\makecell[c]{54.4 \\ \color{pink}53.9}  & 
\makecell[c]{0.358 \\ \color{pink}0.355}\\ \hline

\textbf{AMIE \cite{galarraga2013amie}}  &
\makecell[c]{\color{velvet}337.0}&
\makecell[c]{\color{velvet}64.6}&
\makecell[c]{\color{velvet}0.370} &
\makecell[c]{\color{velvet}309.7}&
\makecell[c]{\color{velvet}{88.1}} & 
\makecell[c]{\color{velvet}0.797}& 
\makecell[c]{\color{velvet}1909}& 
\makecell[c]{\color{velvet}36.2}&
\makecell[c]{\color{velvet}0.201} &
\makecell[c]{\color{velvet}1872} &
\makecell[c]{\color{velvet}{47.7}}&
\makecell[c]{\color{velvet}0.308} \\ \hline

\multicolumn{13}{|p{478pt}|}{

	\textcolor{black}{$\bullet$ Published results}
	\textcolor{blue}{$\bullet$ OpenKE (\url{https://github.com/thunlp/OpenKE )}}
	\textcolor{green}{$\bullet$ ComplEx (\url{https://github.com/ttrouill/complex )}}
	\textcolor{orange}{$\bullet$ ConvE (\url{https://github.com/TimDettmers/ConvE )}}\newline
	\textcolor{cyan}{$\bullet$ RotatE (\url{https://github.com/DeepGraphLearning/KnowledgeGraphEmbedding )}}
	\textcolor{pink}{$\bullet$ TuckER (\url{https://github.com/ibalazevic/TuckER )}}
	\textcolor{velvet}{$\bullet$ AMIE (produced by us)}
	}\\\hline
\end{tabular}
\end{table*}
\egroup

\bgroup
\begin{table*}
\caption{\small Link prediction results on WN18 and WN18RR}\label{table:results wn}
\scriptsize
  \setlength{\tabcolsep}{2pt}
  \begin{tabular}{|P{68pt}|P{24pt}|P{38pt}|P{25pt}|P{25pt}|P{42pt}|P{29pt}||P{24pt}|P{38pt}|P{25pt}|P{25pt}|P{42pt}|P{29pt}|}
  \hline
    
\multicolumn{7}{|c||}{\textbf{\small WN18}}&\multicolumn{6}{c|}{\textbf{\small WN18RR}}  \\\hline
&\multicolumn{3}{c|}{\textbf{\small Raw measures}}&\multicolumn{3}{c||}{\textbf{\small Filtered measures}}&\multicolumn{3}{c|}{\textbf{\small Raw measures}}&\multicolumn{3}{c|}{\textbf{\small Filtered measures}}\\\hline
 \textbf{\small Model} & \textbf{\mr} & \textbf{\hitten} & \textbf{\mrr} & \textbf\fmr& \textbf\fhitten&\textbf\fmrr &  \textbf\mr & \textbf\hitten &\textbf\mrr& \textbf\fmr & \textbf\fhitten &\textbf\fmrr    \\ \hline

\textbf{TransE \cite{bordes2013translating}}   & 
\makecell[c]{263.0 \\ \color{blue}142.4} & 
\makecell[c]{75.4 \\ \color{blue}75.4} & 
\makecell[c]{-- \\ \color{blue}0.395} & 
\makecell[c]{251.0\\ \color{blue}130.8 } &
\makecell[c]{89.2 \\\color{blue}86.0} & 
\makecell[c]{--\\ \color{blue}0.521} &

\makecell[c]{--\\ \color{blue}2414.7} & 
\makecell[c]{--\\ \color{blue}47.2} &  
\makecell[c]{--\\\color{blue}0.176} & 
\makecell[c]{--\\ \color{blue}2401.3} &
\makecell[c]{--\\\color{blue}51.0} &
\makecell[c]{--\\\color{blue}0.224}\\ \hline

\textbf{TransH \cite{TransH}} & 
\makecell[c]{318.0  \\\color{blue}190.1} & 
\makecell[c]{75.4\\\color{blue}76.2} &
\makecell[c]{--\\\color{blue}0.434}&
\makecell[c]{303.0\\ \color{blue}178.7} & 
\makecell[c]{86.7\\\color{blue}86.1} &
\makecell[c]{--\\\color{blue}0.570}&

\makecell[c]{--\\\color{blue}2616} & 
\makecell[c]{--\\\color{blue}46.9}& 
\makecell[c]{--\\\color{blue}0.178}& 
\makecell[c]{--\\\color{blue}2602}&
\makecell[c]{--\\\color{blue}50.4}&  
\makecell[c]{--\\\color{blue}0.224}\\ \hline

\textbf{TransR \cite{TransR}}  &
\makecell[c]{232.0 \\\color{blue}199.7} &
\makecell[c]{78.3\\\color{blue}77.8}&
\makecell[c]{--\\\color{blue}0.441 }&
\makecell[c]{219.0\\  \color{blue}187.9} & 
\makecell[c]{91.7\\ \color{blue}87.3}&
\makecell[c]{--\\  \color{blue}0.583}&

\makecell[c]{--\\ \color{blue}2847} &
\makecell[c]{--\\\color{blue}48.1 } & 
\makecell[c]{--\\ \color{blue}0.184} &
\makecell[c]{-- \\ \color{blue}2834}  &
\makecell[c]{--\\ \color{blue}51.0} &
\makecell[c]{--\\ \color{blue}0.235} \\ \hline

\textbf{TransD \cite{TransD}}  &
\makecell[c]{242.0\\ \color{blue}202.5} & 
\makecell[c]{79.2 \\ \color{blue}79.5} & 
\makecell[c]{-- \\ \color{blue} 0.421}& 
\makecell[c]{229\\  \color{blue}190.6}&
\makecell[c]{92.5\\ \color{blue}91.0} &
\makecell[c]{--\\ \color{blue}0.569}&

\makecell[c]{--\\ \color{blue}2967}&
\makecell[c]{--\\ \color{blue}47.4} &
\makecell[c]{--\\ \color{blue}0.172}  &
\makecell[c]{--\\ \color{blue}2954} &
\makecell[c]{--\\ \color{blue}50.6} &
\makecell[c]{--\\ \color{blue}0.219}\\ \hline



 \textbf{DistMult \cite{DistMult} } &
 \makecell[c]{-- \\ \color{blue}452.9 \\ \color{green}915.0} & 
 \makecell[c]{--\\ \color{blue}80.9\\ \color{green}80.7 } &
 \makecell[c]{-- \\ \color{blue}0.531\\ \color{green} 0.558}& 
 \makecell[c]{-- \\ \color{blue}438.5\\ \color{green} 902.1}  &
 \makecell[c]{94.2\\ \color{blue}93.9\\  \color{green} 93.5}  & 
 \makecell[c]{0.83 \\  \color{blue}0.759\\  \color{green} 0.834} & 
 
 \makecell[c]{-- \\ \color{blue}3798.1\\ \color{green}--} & 
 \makecell[c]{-- \\ \color{blue}46.2\\  \color{green}--}& 
 \makecell[c]{-- \\ \color{blue}0.264\\ \color{green}-- }&
 \makecell[c]{--\\ \color{blue}3784\\ \color{green}-- }&
 \makecell[c]{-- \\ \color{blue}47.5\\ \color{green}-- } &
 \makecell[c]{-- \\ \color{blue}0.371\\ \color{green}-- }\\ \hline

\textbf{ComplEx \cite{complex}}  &
\makecell[c]{-- \\ \color{blue}477.3\\ \color{green} 636.1 } &
\makecell[c]{--\\ \color{blue}81.5\\ \color{green}80.5} &
\makecell[c]{0.587 \\ \color{blue}0.597\\ \color{green}0.584} & 
\makecell[c]{--\\ \color{blue}462.7\\ \color{green}622.7}  &
\makecell[c]{94.7\\ \color{blue}94.6\\ \color{green}94.5 }&
\makecell[c]{0.941\\  \color{blue}0.902\\ \color{green}0.940}&

\makecell[c]{--\\ \color{blue}3755.9\\ \color{green}-- }&
\makecell[c]{--\\ \color{blue}46.7\\ \color{green}--}&
\makecell[c]{-- \\ \color{blue}0.276\\ \color{green}-- } &
\makecell[c]{--\\ \color{blue}3741.7\\ \color{green}--} & 
\makecell[c]{--\\ \color{blue}47.4\\ \color{green}--} & 
\makecell[c]{--\\ \color{blue}0.398\\ \color{green}--}\\ \hline 

\textbf{ConvE \cite{ConvE}}  &
\makecell[c]{--\\ \color{orange}413.1}  &
\makecell[c]{--\\\color{orange}80.6}& 
\makecell[c]{--\\\color{orange}0.574} &
\makecell[c]{504\\\color{orange}396.6} &
\makecell[c]{95.5\\\color{orange}95.5} &
\makecell[c]{0.94\\\color{orange}0.945}&

\makecell[c]{--\\\color{orange}5007.3}   & 
\makecell[c]{-- \\\color{orange}47.9}  &
\makecell[c]{--\\\color{orange}0.261} &
\makecell[c]{5277 \\ \color{orange}4992.7}&
\makecell[c]{48.0 \\ \color{orange}50.4}  & 
\makecell[c]{0.46 \\ \color{orange}0.429}\\ \hline

\textbf{RotatE \cite{RotatE}}  &
\makecell[c]{--\\ \color{cyan}286.2}  &
\makecell[c]{--\\\color{cyan}81.1}& 
\makecell[c]{--\\\color{cyan}0.584} &
\makecell[c]{309\\\color{cyan}269.7} &
\makecell[c]{95.9\\\color{cyan}96.0} &
\makecell[c]{0.949\\\color{cyan}0.950}&

\makecell[c]{--\\\color{cyan}3374}   & 
\makecell[c]{-- \\\color{cyan}53.0}  &
\makecell[c]{--\\\color{cyan}0.306} &
\makecell[c]{3340 \\ \color{cyan}3359.8}&
\makecell[c]{57.1 \\ \color{cyan}57.3}  & 
\makecell[c]{0.476 \\ \color{cyan}0.476}\\ \hline

\textbf{TuckER \cite{TuckER}}  &
\makecell[c]{--\\ \color{pink}484.7}  &
\makecell[c]{--\\\color{pink}80.6}& 
\makecell[c]{--\\\color{pink}0.576} &
\makecell[c]{--\\\color{pink}468.1} &
\makecell[c]{95.8\\\color{pink}95.8} &
\makecell[c]{0.953\\\color{pink}0.950}&

\makecell[c]{--\\\color{pink}6598}   & 
\makecell[c]{-- \\\color{pink}46.8}  &
\makecell[c]{--\\\color{pink}0.272} &
\makecell[c]{ --\\ \color{pink}6584}&
\makecell[c]{52.6 \\ \color{pink}50.2}  & 
\makecell[c]{0.470 \\ \color{pink}0.451}\\ \hline

\textbf{AMIE \cite{galarraga2013amie}}  &
\makecell[c]{\color{velvet}1299.8}&
\makecell[c]{\color{velvet}94.0}&
\makecell[c]{\color{velvet}0.931} &
\makecell[c]{\color{velvet}1299.1}&
\makecell[c]{\color{velvet}{94.0}} & 
\makecell[c]{\color{velvet}0.940}& 
\makecell[c]{\color{velvet}12963}& 
\makecell[c]{\color{velvet}35.6}&
\makecell[c]{\color{velvet}0.357} &
\makecell[c]{\color{velvet}12957} &
\makecell[c]{\color{velvet}{35.6}}&
\makecell[c]{\color{velvet}0.357} \\ \hline
\multicolumn{13}{|p{478pt}|}{

	\textcolor{black}{$\bullet$ Published results}
	\textcolor{blue}{$\bullet$ OpenKE (\url{https://github.com/thunlp/OpenKE )}}
	\textcolor{green}{$\bullet$ ComplEx (\url{https://github.com/ttrouill/complex )}}
	\textcolor{orange}{$\bullet$ ConvE (\url{https://github.com/TimDettmers/ConvE )}}\newline
	\textcolor{cyan}{$\bullet$ RotatE (\url{https://github.com/DeepGraphLearning/KnowledgeGraphEmbedding )}}
	\textcolor{pink}{$\bullet$ TuckER (\url{https://github.com/ibalazevic/TuckER )}}
	\textcolor{velvet}{$\bullet$ AMIE (produced by us)}
	}\\\hline

\end{tabular}
\end{table*}
\egroup

(4) The performance measures in Section~\ref{sec:eval-measure} are based on the closed-world assumption and thus have flaws when a model correctly predicts a triple that does not exist in the ground-truth dataset. More specifically, if a corrupted triple of a given test triple does not exist in the training or test set, it is considered incorrect. However, the corrupted triple might be correct as well. If a model ranks the corrupted triple higher than the test triple itself, its accuracy measures will be penalized, which contradicts with the exact goal of link prediction---finding correct triples that do not already exist in the dataset. While this defect of the accuracy measures is applicable on all types of relation,~\footnote{Fundamentally it is because the models are evaluated by ranking instead of binary classification of triples.} it is more apparent in evaluating a method that is capable of leveraging the characteristics of Cartesian product relations. Such a method would mark many triples non-existent in the dataset as correct and further rank many of them higher than the test triples. 

Table~\ref{table:cartesian} uses several measures to show the accuracy of the prediction method based on the Cartesian product property, as explained in (2).~\footnote{For coping with space limitations, we shortened the names of some measures, e.g., \fhitten is shortened as \fhten.} The method's accuracy is evaluated using both FB15k and the larger Freebase snapshot as the ground truth. The table also shows the results of using TransE. In all cases FB15k training set is used as the training data for making predictions. The table presents the results on all 9 Cartesian product relations we detected from FB15k, which are listed in Table~\ref{table:cartesian-rel}. Some of them are detected as Cartesian product relations by applying the aforementioned process over the training set of FB15k and some are detected over the Freebase snapshot. We can make several observations regarding the results in Table~\ref{table:cartesian}.  First, the performance of using Cartesian product property is higher when Freebase instead of FB15k is the ground truth. This is because Freebase subsumes FB15k and thus is affected less by the defect mentioned in (4). Second, using Cartesian product property is more accurate than embedding models such as TransE, especially when the Freebase snapshot is used as the ground truth to calculate filtered measures. (Note that using Freebase as the ground truth will not affect unfiltered measures such as \mr. Therefore we do not repeat those measures in the table.) 
For example, consider predicting triples in relation $\rel{r2}$. We observed that the Cartesian product property attained a \fmrr\ of 0.80 using FB15k as ground truth, in comparison to 0.42 by TransE. The accuracy is further improved to 1 when using the Freebase snapshot as the ground truth.

%% file: sec-create.tex
\subsection{Identifying the Most Probable Freebase Snapshot Used for Producing FB15k}\label{sec:snapshot}

In order to understand the various defects in FB15k and their root cause, we searched for the same Freebase snapshot that was used to create FB15k. When it was active, Freebase maintained periodic snapshots, more frequent than monthly. It is unclear from~\cite{bordes2013translating} which snapshot was used to create FB15k.\footnote{We had email communication with the authors of~\cite{bordes2013translating}. They could not remember the exact date of the Freebase snapshot used for producing FB15k.}  To derive the most probable timestamp of the snapshot from which FB15k was produced, we compared the snapshots available at {\small\url{https://commondatastorage.googleapis.com/freebase-public/}} as of June 2019. Particularly, we considered the snapshots in the several months before~\cite{bordes2013translating} was published. We analyzed to what degree these snapshots overlap with FB15k. 

The May 5, 2013 snapshot (i.e., {\small\url{https://commondatastorage.googleapis.com/freebase-public/rdf/freebase-rdf-2013-05-05-00-00.gz}}) has the largest overlap among these snapshots, as it contains 99.54\% of the triples in FB15k. We thus concluded that FB15k was most likely drawn from a snapshot around May 5, 2013, which can be approximated by the snapshot on that exact date. 
 
\begin{figure}
    \centering
    \includegraphics[trim={2.7cm 12cm 3.7cm 3.5cm},clip,width=0.48\textwidth]{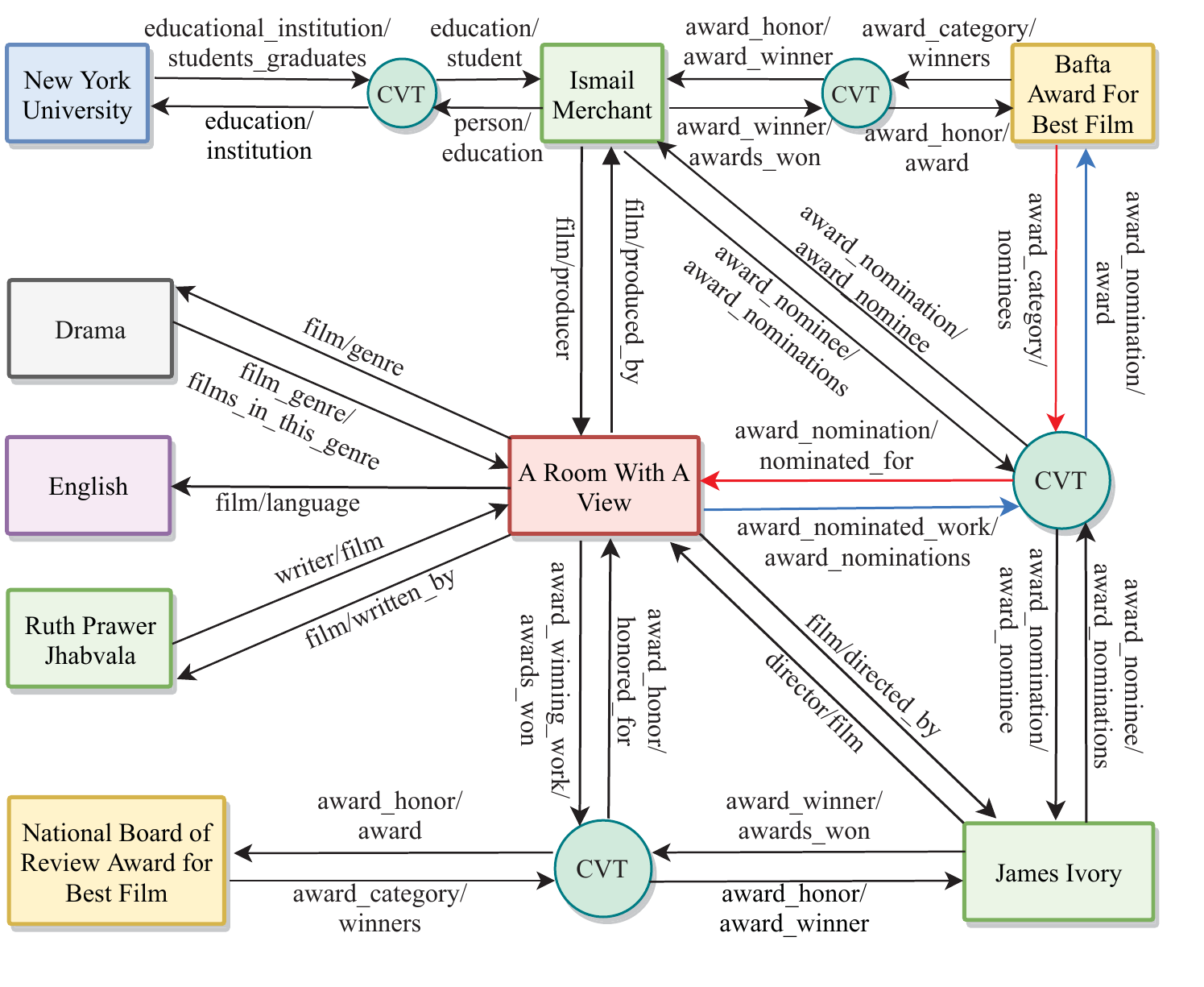}
    \caption{\small Mediator (CVT) nodes in Freebase}
    \label{fig:mid node}
\end{figure}

\emph{Mediator nodes}, also called \emph{compound value type} (CVT) nodes, are used in Freebase to represent multiary relationships. For example, Figure~\ref{fig:mid node} shows several CVT nodes. The rightmost CVT node is connected to an \entity{award}, two \entity{nominee} nodes and a \entity{work} through various relations. In the May 2013 Freebase snapshot, for many (but not all) CVT nodes, additional \emph{concatenated edges} were created. Specifically, for such a CVT node, multiple triples were created, each a concatenation of two edges connected through the CVT node. These binary relationships partly capture the multiary relationship represented by the CVT node. 
For instance, the triples \triple{Bafta Award For Best Film}{award\_category/nominees}{CVT} and \triple{CVT}{award\_nomination/nominated\_for}{A Room With A View} in Figure~\ref{fig:mid node} would be concatenated to form a triple between the award and the work nominated for the award.
The concatenation of two relations $\rel{r}_1$ and $\rel{r}_2$ is written as $\rel{r}_1.\rel{r}_2$. There are 54,541,700 concatenated triples in the May 2013 snapshot.

{\flushleft \textbf{Concatenated Edges}}\hspace{2mm}

FB15k does not include any CVT nodes or edges adjacent to such nodes from Freebase. However, it kept most concatenated edges (or maybe all, although we have no absolute way to verify, since nowhere we can find all Freebase snapshots that have ever been created). All the 707 concatenated relations in FB15k exist in the May 2013 snapshot. Among the 592,213 triples in FB15k, 396,148 are concatenated and 394,947 of them could be found in the snapshot.

%% file: sec-exp.tex
\section{Experiments}\label{sec:exp}

\begin{table}
\caption{\small Percentages of test triples, among those on which various models outperformed TransE, that have reverse and duplicate triples in training set}
\label{table:reverse_triple_percentage}
\centering
\small
\setlength{\tabcolsep}{2pt}
\renewcommand{\arraystretch}{0.8}
\begin{tabular}{|c|c|c|c|c|c|c|} \hline
\multicolumn{5}{|c|}{\textbf{FB15k}}\\\hline
\textbf{\small Model}&\textbf{\fmr}&\textbf{\fhitten} & \textbf{\fhitone}&\textbf{\fmrr}\\ \hline
\textbf{DistMult}& 82.17 \%& 90.78 \%& 95.16 \%& 85.88 \%\\ \hline
\textbf{ComplEx}& 81.24 \%& 90.14 \%& 94.98 \%& 84.67 \%\\ \hline
\textbf{ConvE}& 78.69 \%& 87.12 \%& 91.1 \%& 78.04 \%\\ \hline
\textbf{RotatE}&78.61\% &88.37\% &94.41\% &78.16\% \\ \hline
\textbf{TuckER}&78.96\%&87.76\% &93.65\% &78.87\% \\ \hline
\end{tabular}

\begin{tabular}{|c|c|c|c|c|c|c|} \hline
\multicolumn{5}{|c|}{\textbf{WN18}}\\\hline
\textbf{\small Model}&\textbf{\fmr}&\textbf{\fhitten} & \textbf{\fhitone}&\textbf{\fmrr}\\ \hline
\textbf{DistMult}& 98.43 \%& 99.4 \%& 98.53 \%& 98.55 \%\\ \hline
\textbf{ComplEx}& 97.93 \%& 98.72 \%& 99.1 \%& 97.73 \%\\ \hline
\textbf{ConvE}& 96.42 \%& 96.7 \%& 98.96 \%& 95.85 \%\\ \hline
\textbf{RotatE}&95.17\% &95.82\% &98.74\% &94.00\% \\ \hline
\textbf{TuckER}&95.75\%&95.18\% &98.45\% &95.70\% \\ \hline
\end{tabular}\vspace{4mm}
\end{table}

\begin{table*}
\caption{\small Number of relations on which each model is the most accurate}\label{table:num fb}
\centering
\scriptsize
\setlength{\tabcolsep}{2pt}
\renewcommand{\arraystretch}{0.8}
\begin{tabular}{|c|c|c|c|c|c|c|c|c|c|c|c|c|} \hline
&\multicolumn{4}{c|}{\small\textbf{FB15-237}}&\multicolumn{4}{c|}{\small\textbf{WN18RR}}&\multicolumn{4}{c|}{\small\textbf{YAGO3-10}}\\\hline
\textbf{\small Model} & \textbf{\fmr} & \textbf{\fhten} & \textbf{\fhone} & \textbf{\fmrr}& \textbf{\fmr} & \textbf{\fhten} & \textbf{\fhone} & \textbf{\fmrr}& \textbf{\fmr} & \textbf{\fhten} & \textbf{\fhone} & \textbf{\fmrr}\\ \hline
\textbf{TransE}  & 17 & 43 & 23 & 19& 8 & 2 & 0 & 1& 9 & 10 & 3 & 3\\\hline
\textbf{DistMult}  & 5 & 23 & 13 & 6& 0 & 1 & 0 & 0& 4 & 3 & 6 & 1\\\hline
\textbf{ComplEx} & 6 & 20 & 12 & 5& 0 & 2& 0 & 0& 6 & 10 & 7 & 4\\\hline
\textbf{ConvE} & 8 & 41 & 30 & 15& 1 & 3 & 3 & 3& 4 & 7 & 7 & 2\\\hline
\textbf{RotatE} & 77 & 92 & 58 & 55 & 2 & 9 & 4 & 5& 2 & 13 & 8 & 5\\\hline
\textbf{TuckER} & 95 & 112 & 101 & 91& 2 & 2 & 6 & 4& 5 & 13 & 10 & 10\\\hline
\textbf{AMIE}  & 31 & 66 & 68 & 49& 1 & 2 & 3 & 3& 4 & 6 & 12 & 9\\\hline
\end{tabular}
\end{table*}

\subsection{FB15k-237, WN18RR, YAGO3-10-DR}\label{sec:newdataset}
Among the first to document the data redundancy in FB15k, Toutanova and Chen~\cite{toutanova2015observed} created FB15k-237 from FB15k by removing such redundancy. They first limited the set of relations in FB15k to the most frequent 401 relations. Their approach of removing redundancy is essentially the same as the equations for detecting duplicate and reverse duplicate relations in Section~\ref{sec:duplicate}, likely with different thresholds. For each pair of such redundant relations, only one was kept. This process decreased the number of relations to 237. They also removed all triples in test and validation sets whose entity pairs were directly linked in the training set through any relation. This step could incorrectly remove useful information. For example, \rel{place\_of\_birth} and \rel{place\_of\_death} may have many overlapping subject-object pairs, but they are not semantically redundant. Furthermore, the creation of FB15k-237 did not resort to the absolutely accurate reverse relation information encoded by \rel{reverse\_property}. Finally, it does not identify Cartesian product relations. 
Nevertheless, we used both FB15k-237 and FB15k in our experiments. This allows us to corroborate the experiment results with those from a number of recent studies.

\begin{figure*}
    \centering
    \includegraphics[trim={7cm 0 5cm 0cm},clip,width=\textwidth]{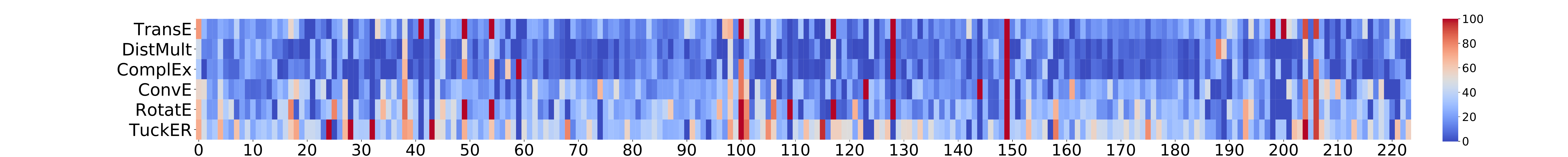}   
    \caption{\small Percentage of triples on which each method outperforms others, separately for each FB15k-237 relation}
    \label{fig:heatmap}
\end{figure*}

\begin{figure}
    \centering
    \includegraphics[width=0.45\textwidth]{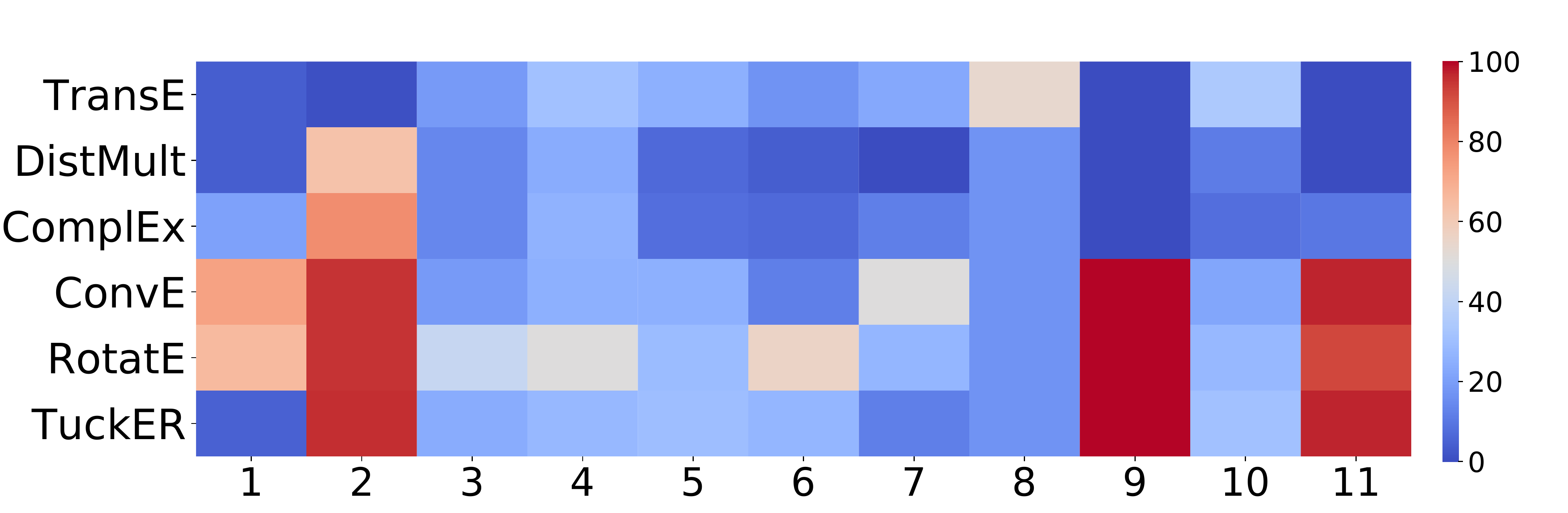}
    \caption{\small Percentage of triples on which each method outperforms others, for each WN18RR relation}
    \label{fig:heatmap_wn18rr}
\end{figure}

To remove the WN18 reverse relations mentioned in Section~\ref{sec:redundancy}, Dettmers et al.~\cite{ConvE} created WN18RR by keeping just one relation from each pair of reverse relations. The resulting dataset WN18RR has 40,943 entities in 11 relations. However, this dataset still contains symmetric (i.e., self-reciprocal) relations---a special case of reverse relations where a relation is the reverse of itself. Particularly, more than 34\% of the training triples in WN18RR belong to the symmetric relation \rel{derivationally\_related\_form} which is for terms in different syntactic categories that have the same morphological root. For instance, both triples \triple{question}{derivationally\_related\_form}{inquire} and \triple{inquire}{derivationally\_related\_form}{question} are in the training set. Among the 11 relations in WN18RR's training set, 3 are self-reciprocal, which account for 30,933 of the 86,835 training triples. 28,835 out of these 30,933 triples form reverse pairs. For the remaining 2,098 triples, 1,052 form reverse pairs of with 1,052 triples (around 33.57\%) of the test set.

As mentioned in Section~\ref{sec:redundancy}, YAGO3-10 has two near-duplicate relations \rel{isAffiliatedTo} and \rel{playsFor} and 3 semantically symmetric relations \rel{hasNeighbor}, \rel{isConnectedTo}, and \rel{isMarriedTo}. We removed \rel{playsFor}. In the training set, for each pair of redundant triples belonging to symmetric relations, we only kept one. Moreover, we removed a triple from the test and validation sets if it belongs to any symmetric relation and its entity pairs are directly linked in the training set. We call the resulting dataset YAGO3-10-DR, of which the statistics can be found in Table~\ref{table:datasets}.

\subsection{Experiment Setup}

Our experiments were conducted on an Intel-based machine with an Intel Xeon E5-2695 processor running at 2.1GHz, Nvidia Geforce GTX1080Ti GPU, and 256 GB RAM. The experiments used source codes of various methods from several places, including the OpenKE~\cite{openke} repository which covers implementations of TransE, TransH, TransR, TransD, RESCAL, DistMult, and ComplEx, as well as the source code releases of ComplEx (which also covers DistMult), ConvE, RotatE, and TuckER. The URLs of these implementations can be found in Table~\ref{table:results}. 

The different models used in our experiments have different hyperparameters. We used the same hyperparameter settings for FB15k, FB15k-237, WN18, WN18RR, and YAGO3-10 that were used by the developers of the source codes. The details can be found in the codes.

We also experimented with rule-based system AMIE~\cite{galarraga2013amie}. AMIE rules were generated by applying the AMIE+ (\url{https://bit.ly/2Vq2OIB}) code released by the authors of~\cite{amieplus} on the training sets of FB15k, FB15k-237, WN18, WN18RR, and YAGO3-10. The parameters of AMIE+ were set in the same way as in~\cite{finegrained}, for all datasets. For any link prediction task \triple{h}{r}{?} or \triple{?}{r}{t}, all the rules that have relation \rel{r} in the rule head are employed. The instantiations of these rules are used to generate the ranked list of results. For example, for test case \triple{Bill Gates}{place\_of\_birth}{?}, the following rule will be employed: $\triple{?a}{places\_lived/location}{?b}\ \Rightarrow \triple{?a}{place\_of\_birth}{?b}$. Then the instantiations of variable \entity{?b} are  used to find the list of predictions. Several rules may generate the same answer entity. It is imperative to combine the confidence of those rules in some way in order to score the answer entities. We ranked the answer entities by the maximum confidence of the rules instantiating them and broke ties by the number of applicable rules~\cite{finegrained}. \gotoTR{Confidence of the rule is defined by the ratio of instantiation of the rule to the instantiation of its body (how often the rule has been true).}

\begin{figure}
\centering
\begin{subfigure}[b]{0.47\textwidth}
  \centering
  \begin{tikzpicture}
    \centering
    \begin{axis}[
        width  = 0.93*\textwidth,
        height  = 0.4*\textwidth,
        ylabel= Number of relations,
    	label style={font=\tiny},
    	enlargelimits=0.05,
    	ybar=2*\pgflinewidth,
    	bar width=.15cm,
    	legend image code/.code={%
      \draw[#1] (0cm,-0.1cm) rectangle (0.1cm,0.1cm);
    }   ,
    	y label style={at={(axis description cs:0.1,.5)}},
    	x tick label style={font=\tiny},
    	y tick label style={font=\tiny},
    	legend style={font=\tiny,at={(1.19,0)},anchor=south east},
    	enlarge x limits=0.15,
    	symbolic x coords={TransE,DistMult,ComplEx,ConvE,RotatE,TuckER, AMIE},
               xtick=data
               ]
    \addplot
    	coordinates {(TransE,5)(DistMult,1) 
    		 (ComplEx,1) (ConvE,1)(RotatE,7)(TuckER,2)(AMIE,2)};
    \addplot
    	coordinates {(TransE,1)(DistMult,0) 
    		 (ComplEx,0) (ConvE,0)(RotatE,13)(TuckER,7)(AMIE,5)};
    \addplot
    	coordinates {(TransE,4)(DistMult,0) 
    		 (ComplEx,2) (ConvE,5)(RotatE,18)(TuckER,29)(AMIE,20)};
    \addplot
    	coordinates {(TransE,9)(DistMult,5) 
    		 (ComplEx,2) (ConvE,9)(RotatE,17)(TuckER,53)(AMIE,22)};
    \legend{$1$-to-$1$, $1$-to-$n$, $n$-to-$1$, $n$-to-$m$}
    \end{axis}
    \end{tikzpicture}
  \caption{\small{Categorizing the relations on which each method has the best result}} \label{fig:best res cat fb}
\end{subfigure}%

\begin{subfigure}[b]{0.47\textwidth}
  \centering
  \begin{tikzpicture}
    \centering
    \begin{axis}[
        width  = 0.93*\textwidth,
        height  = 0.4*\textwidth,
        x tick label style={
    		/pgf/number format/1000 sep=},
    	ylabel= \%,
    	label style={font=\tiny},
    	enlargelimits=0.05,
    	ybar=2*\pgflinewidth,
    	y label style={at={(axis description cs:0.1,.5)}},
    	x tick label style={font=\tiny},
    	y tick label style={font=\tiny},
    	bar width=.18cm,
    	legend image code/.code={%
      \draw[#1] (0cm,-0.1cm) rectangle (0.1cm,0.1cm);
    }   ,
    	legend style={font=\tiny,at={(1.21,0)},anchor=south east},
    	enlarge x limits=0.15,
    	symbolic x coords={$1$-to-$1$,$1$-to-$n$,$n$-to-$1$,$n$-to-$m$},
               xtick=data
               ]
    \addplot
    	coordinates {($1$-to-$1$,38.46) ($1$-to-$n$,3.85)
    		 ($n$-to-$1$,5.26) ($n$-to-$m$,8.26)};
   \addplot 
    	coordinates {($1$-to-$1$,7.69) ($1$-to-$n$,0)
    		 ($n$-to-$1$,0) ($n$-to-$m$,4.59)};
    \addplot 
    	coordinates {($1$-to-$1$,7.69) ($1$-to-$n$,0)
    		 ($n$-to-$1$,2.63) ($n$-to-$m$,1.83)};
    
     \addplot 
    	coordinates {($1$-to-$1$,7.69) ($1$-to-$n$,0)
    		 ($n$-to-$1$,6.58) ($n$-to-$m$,8.26)};		 
    \addplot 
    	coordinates {($1$-to-$1$,53.85) ($1$-to-$n$,50)
    		 ($n$-to-$1$,23.68) ($n$-to-$m$,15.6)};
	 \addplot 
        coordinates {($1$-to-$1$,15.38) ($1$-to-$n$,26.92)
    		 ($n$-to-$1$,38.16) ($n$-to-$m$,48.62) };
    \addplot [fill=red]
        coordinates {($1$-to-$1$,15.38) ($1$-to-$n$,19.23)
    		 ($n$-to-$1$,26.32) ($n$-to-$m$,20.18) };
    \legend{TransE,DistMult,ComplEx,ConvE, RotatE, TuckER, AMIE}
    \end{axis}
    \end{tikzpicture}
  \caption{\small{Break-down of methods achieving best performance on each type of relations}}\label{fig:best res cat percentage fb}
\end{subfigure}%
\caption{\small Models with best \fmrr\ on FB15k-237}
\label{fig:res_by_rel_type_fb}
\end{figure}
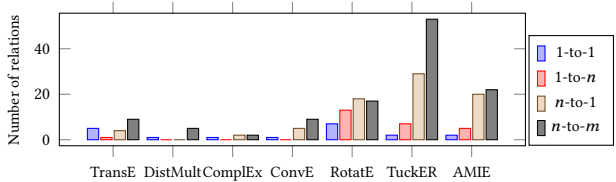
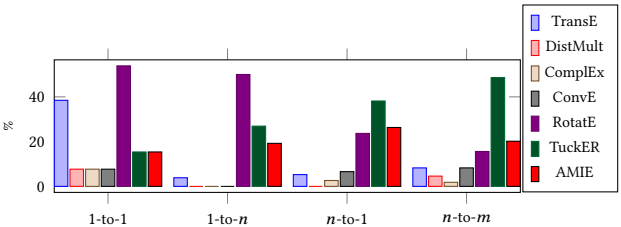

\begin{table}[t]
\caption{\small \fhitten\ by category of relations on FB15k-237}\label{table:amieh10_fb}
\centering
\scriptsize
\setlength{\tabcolsep}{2pt}
\begin{tabular}{|c|c|c|c|c|c|c|c|c|} \hline
 &\multicolumn{2}{c|}{\textbf{$1$-to-$1$}}&\multicolumn{2}{c|}{\textbf{$1$-to-$n$}}&\multicolumn{2}{c|}{\textbf{$n$-to-$1$}}&\multicolumn{2}{c|}{\textbf{$n$-to-$m$}}\\\hline
\textbf{Model}  & \makecell[c]{\textbf{Left}\\ \textbf{\tfhten}} & \makecell[c]{\textbf{Right}\\ \textbf{\tfhten}}& \makecell[c]{\textbf{Left}\\ \textbf{\tfhten}} & \makecell[c]{\textbf{Right}\\ \textbf{\tfhten}}& \makecell[c]{\textbf{Left}\\ \textbf{\tfhten}} & \makecell[c]{\textbf{Right} \\\textbf{\tfhten}}& \makecell[c]{\textbf{Left}\\ \textbf{\tfhten}} & \makecell[c]{\textbf{Right}\\ \textbf{\tfhten}}\\ \hline
\textbf{TransE} & \underline{55.21} & \underline{55.21} & 60.40 & 9.59 & 11.74 & 84.60 & 40.52 & 56.12\\\hline
\textbf{DistMult} & 47.92 & 46.35 & 43.62 & 3.63 & 4.34 & 76.06 & 36.09 & 51.48\\\hline
\textbf{ComplEx}  & 47.40& 46.88 & 36.66 & 3.94 & 5.13 & 75.82& 36.95 & 52.48\\\hline
\textbf{ConvE}    & 27.60 &  26.56 &59.63 & 11.29 &14.38 & 85.34 & 41.20 & 56.73\\\hline
\textbf{RotatE}   & \textbf{59.38}  &  \textbf{58.85}  & \textbf{67.52} & \underline{13.61} &17.08 & \underline{87.49} & \underline{47.38} &  \underline{61.49}\\\hline
\textbf{TuckER}   & \underline{55.21}  &  52.60  & \underline{65.58} & \textbf{15.62} & \textbf{21.52} & \textbf{87.84} & \textbf{48.14} &  \textbf{61.54}\\\hline
\textbf{AMIE}     & 43.23  & 44.27  & 46.17 &13.07 &  \underline{18.20} & 80.49 & 42.45 & 55.19 \\ \hline
\end{tabular}
\end{table}

\begin{table}[t]
\caption{\small \fhitten\ by category of relations on WN18RR}\label{table:amie_hit10_wn}
\centering
\scriptsize
\setlength{\tabcolsep}{2pt}
\begin{tabular}{|c|c|c|c|c|c|c|c|c|} \hline
 &\multicolumn{2}{c|}{\textbf{$1$-to-$1$}}&\multicolumn{2}{c|}{\textbf{$1$-to-$n$}}&\multicolumn{2}{c|}{\textbf{$n$-to-$1$}}&\multicolumn{2}{c|}{\textbf{$n$-to-$m$}}\\\hline
\textbf{Model}  & \makecell[c]{\textbf{Left}\\ \textbf{\tfhten}} & \makecell[c]{\textbf{Right}\\ \textbf{\tfhten}}& \makecell[c]{\textbf{Left}\\ \textbf{\tfhten}} & \makecell[c]{\textbf{Right}\\ \textbf{\tfhten}}& \makecell[c]{\textbf{Left}\\ \textbf{\tfhten}} & \makecell[c]{\textbf{Right} \\\textbf{\tfhten}}& \makecell[c]{\textbf{Left}\\ \textbf{\tfhten}} & \makecell[c]{\textbf{Right}\\ \textbf{\tfhten}}\\ \hline
\textbf{TransE}   & \underline{92.86} & \underline{92.86} &  42.32 &    15.79 &    14.32  &   \underline{34.90}   &   92.92 &    93.72\\\hline
\textbf{DistMult} & \textbf{97.62} & \underline{92.86} &  24.21 &    6.32  &    5.78  &    33.83 &    95.75  &   \textbf{95.75}\\\hline
\textbf{ComplEx}  & \textbf{97.62} & \textbf{97.62} &  29.47 &    7.58  &    5.72  &    33.02 &    \underline{95.84} &    95.31 \\\hline
\textbf{ConvE}    & \textbf{97.62} & \textbf{97.62} &  \underline{44.63} &    16.84 &    11.57  &   31.34  &   95.22  &   94.96 \\\hline
\textbf{RotatE}   & \textbf{97.62} & \textbf{97.62} &  \textbf{53.68} &    \textbf{28.84} &    \textbf{20.98} &    \textbf{43.04}  &   \textbf{96.11}  &   \underline{95.58}\\\hline
\textbf{TuckER}   & \textbf{97.62} & \textbf{97.62} &  40.21 &    \underline{18.95} &    \underline{15.13}  &   30.26 &    94.96  &   91.77\\\hline
\textbf{AMIE}     & \textbf{97.62} & \textbf{97.62} &  1.26  &    1.26  &   1.41 &     1.41  &    92.83  &   92.83 \\ \hline
\end{tabular}
\end{table}


\bgroup
\begin{table}
\caption{\small Link prediction results on YAGO3-10}\label{table:results yago}
\scriptsize
  \setlength{\tabcolsep}{2pt}
  \begin{tabular}{|c|c|c|c|c||c|c|c|c|}
  \hline
    
\multicolumn{5}{|c||}{\textbf{\small YAGO3-10}} & \multicolumn{4}{c|}{\textbf{\small YAGO3-10-DR}}\\\hline
 \textbf{\small Model} & \textbf{\tfhone} & \textbf\tfmr& \textbf\tfhten & \textbf{\tfmrr}& \textbf{\tfhone} & \textbf\tfmr& \textbf\tfhten & \textbf{\tfmrr} \\ \hline

\textbf{TransE}   & 
\makecell[c]{--\\\color{cyan}40.7} & 
\makecell[c]{--\\ \color{cyan}1193.4} &
\makecell[c]{--\\\color{cyan}67.9} &
\makecell[c]{--\\\color{cyan}0.504}&

\makecell[c]{--\\\color{cyan}11.7} &
\makecell[c]{-- \\ \color{cyan}2355.9}&
\makecell[c]{-- \\ \color{cyan}32.3}  & 
\makecell[c]{-- \\ \color{cyan}0.19}\\ \hline

 \textbf{DistMult} &
 \makecell[c]{--\\\color{cyan}34.2 \\ \color{orange}42.4}& 
 \makecell[c]{--\\ \color{cyan}1712.8 \\ \color{orange}2685.1}&
 \makecell[c]{--\\ \color{cyan}64.9 \\ \color{orange}66.0} &
 \makecell[c]{--\\ \color{cyan}0.448 \\ \color{orange}0.51}&
 
 \makecell[c]{--\\\color{cyan}9.6 \\ \color{orange}13.3}& 
 \makecell[c]{--\\ \color{cyan}7509.3 \\ \color{orange}5553.2}&
 \makecell[c]{--\\ \color{cyan}28.8 \\ \color{orange}30.7} &
 \makecell[c]{--\\ \color{cyan}0.161 \\ \color{orange}0.192}\\ \hline

\textbf{ComplEx}  &
\makecell[c]{--\\ \color{cyan}35.5 \\ \color{orange}44.9} &
\makecell[c]{--\\ \color{cyan}3076.4\\ \color{orange}2911.7} & 
\makecell[c]{--\\ \color{cyan}64.6\\ \color{orange}67.8} & 
\makecell[c]{--\\ \color{cyan}0.455\\ \color{orange}0.53}& 

\makecell[c]{--\\ \color{cyan}9.7 \\ \color{orange}14.3} &
\makecell[c]{--\\ \color{cyan}8498.1\\ \color{orange}6077} & 
\makecell[c]{--\\ \color{cyan}28.8\\ \color{orange}31.5} & 
\makecell[c]{--\\ \color{cyan}0.162\\ \color{orange}0.201}\\ \hline

\textbf{ConvE}  &
\makecell[c]{45.0\\\color{orange}46.2} &
\makecell[c]{2792 \\ \color{orange}1598.8}&
\makecell[c]{66.0 \\ \color{orange}68.6}  & 
\makecell[c]{0.52 \\ \color{orange} 0.542}&

\makecell[c]{--\\\color{orange}14.7} &
\makecell[c]{--\\ \color{orange}4453.3}&
\makecell[c]{-- \\ \color{orange}31.5}  & 
\makecell[c]{-- \\ \color{orange}0.204}\\ \hline

\textbf{RotatE}  &
\makecell[c]{40.2\\\color{cyan}40.5} &
\makecell[c]{1767 \\ \color{cyan}1809.4}&
\makecell[c]{67.0 \\ \color{cyan}67.4}  & 
\makecell[c]{0.495 \\ \color{cyan}0.499}&

\makecell[c]{--\\\color{cyan}15.3} &
\makecell[c]{-- \\ \color{cyan}3084.2}&
\makecell[c]{-- \\ \color{cyan}33.2}  & 
\makecell[c]{-- \\ \color{cyan}0.214}\\ \hline

\textbf{TuckER }  &
\makecell[c]{--\\\color{pink}40.7} &
\makecell[c]{--\\ \color{pink}2293.8}&
\makecell[c]{-- \\ \color{pink}66.1}  & 
\makecell[c]{-- \\ \color{pink}0.496}&

\makecell[c]{--\\\color{pink}14.8} &
\makecell[c]{--\\ \color{pink}6068.8}&
\makecell[c]{-- \\ \color{pink}32}  & 
\makecell[c]{-- \\ \color{pink}0.207}\\ \hline

\textbf{AMIE}  &
\makecell[c]{\color{velvet}55.8} &
\makecell[c]{\color{velvet}24133}&
\makecell[c]{\color{velvet}57.96} & 
\makecell[c]{\color{velvet}0.565} &

\makecell[c]{\color{velvet}--} &
\makecell[c]{\color{velvet}--}&
\makecell[c]{\color{velvet}--} & 
\makecell[c]{\color{velvet}--} \\ \hline

\end{tabular}
\end{table}
\egroup


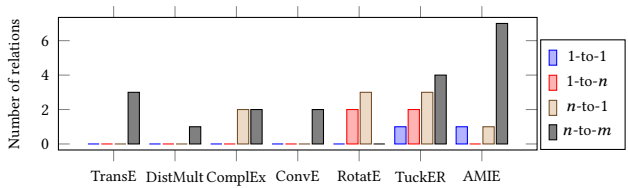
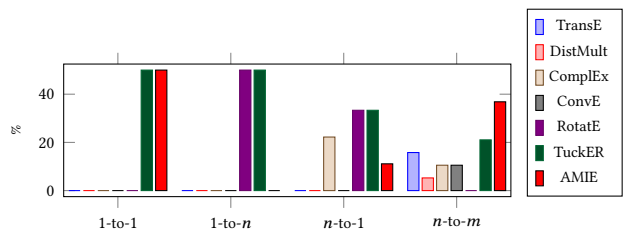
\begin{figure}
\centering
\begin{subfigure}[b]{0.47\textwidth}
  \centering
  \begin{tikzpicture}
    \centering
    \begin{axis}[
        width  = 0.95*\textwidth,
        height  = 0.4*\textwidth,
        ylabel= Number of relations,
    	label style={font=\tiny},
    	enlargelimits=0.05,
    	ybar=2*\pgflinewidth,
    	bar width=.15cm,
    	legend image code/.code={%
      \draw[#1] (0cm,-0.1cm) rectangle (0.1cm,0.1cm);
    }   ,
    	y label style={at={(axis description cs:0.1,.5)}},
    	x tick label style={font=\tiny},
    	y tick label style={font=\tiny},
    	legend style={font=\tiny,at={(1.185,0)},anchor=south east},
    	enlarge x limits=0.15,
    	symbolic x coords={TransE,DistMult,ComplEx,ConvE,RotatE,TuckER, AMIE},
               xtick=data
               ]
    \addplot
    	coordinates {(TransE,0)(DistMult,0) 
    		 (ComplEx,0) (ConvE,0)(RotatE,0)(TuckER,1)(AMIE,1)};
    \addplot
    	coordinates {(TransE,0)(DistMult,0) 
    		 (ComplEx,0) (ConvE,0)(RotatE,2)(TuckER,2)(AMIE,0)};
    \addplot
    	coordinates {(TransE,0)(DistMult,0) 
    		 (ComplEx,2) (ConvE,0)(RotatE,3)(TuckER,3)(AMIE,1)};
    \addplot
    	coordinates {(TransE,3)(DistMult,1) 
    		 (ComplEx,2) (ConvE,2)(RotatE,0)(TuckER,4)(AMIE,7)};
    \legend{$1$-to-$1$, $1$-to-$n$, $n$-to-$1$, $n$-to-$m$}
    \end{axis}
    \end{tikzpicture}
  \caption{\small{Categorizing the relations on which each method has the best result}} \label{fig:best res cat yago}
\end{subfigure}

\begin{subfigure}[b]{0.47\textwidth}
  \centering
  \begin{tikzpicture}
    \centering
    \begin{axis}[
        width  = 0.9*\textwidth,
        height  = 0.4*\textwidth,
        x tick label style={
    		/pgf/number format/1000 sep=},
    	ylabel= \%,
    	label style={font=\tiny},
    	enlargelimits=0.05,
    	ybar=3*\pgflinewidth,
    	y label style={at={(axis description cs:0.1,.5)}},
    	x tick label style={font=\tiny},
    	y tick label style={font=\tiny},
    	bar width=.15cm,
    	legend image code/.code={%
      \draw[#1] (0cm,-0.1cm) rectangle (0.1cm,0.1cm);
    }   ,
    	legend style={font=\tiny,at={(1.25,0)},anchor=south east},
    	enlarge x limits=0.16,
    	symbolic x coords={$1$-to-$1$,$1$-to-$n$,$n$-to-$1$,$n$-to-$m$},
               xtick=data
               ]
    \addplot 
    	coordinates {($1$-to-$1$,0) ($1$-to-$n$,0)
    		 ($n$-to-$1$,0) ($n$-to-$m$,15.79)};
    \addplot 
    	coordinates {($1$-to-$1$,0) ($1$-to-$n$,0)
    		 ($n$-to-$1$,0) ($n$-to-$m$,5.26)};
    \addplot 
    	coordinates {($1$-to-$1$,0) ($1$-to-$n$,0)
    		 ($n$-to-$1$,22.22) ($n$-to-$m$,10.53)};
    \addplot 
    	coordinates {($1$-to-$1$,0) ($1$-to-$n$,0)
    		 ($n$-to-$1$,0) ($n$-to-$m$,10.53)};
    \addplot
    	coordinates {($1$-to-$1$,0) ($1$-to-$n$,50)
    		 ($n$-to-$1$,33.33) ($n$-to-$m$,0)};
    \addplot 
    	coordinates {($1$-to-$1$,50) ($1$-to-$n$,50)
    		 ($n$-to-$1$,33.33) ($n$-to-$m$,21.05)};
    \addplot[fill=red] 
    	coordinates {($1$-to-$1$,50) ($1$-to-$n$,0)
    		 ($n$-to-$1$,11.11) ($n$-to-$m$,36.84)};
    \legend{TransE,DistMult,ComplEx,ConvE,RotatE,TuckER,AMIE}
    \end{axis}
    \end{tikzpicture}
  \caption{\small{Break-down of methods achieving best performance on each type of relations}}\label{fig:best res cat percentage yago}
\end{subfigure}%
\caption{\small Models with best \fmrr\ on YAGO3-10}
\label{fig:res_by_rel_type_yago}
\end{figure}


\begin{table}[t]
\caption{\small \fhitten\ by category of relations on YAGO31-0}\label{table:amieh10_yago}
\centering
\scriptsize
\setlength{\tabcolsep}{2pt}
\begin{tabular}{|c|c|c|c|c|c|c|c|c|} \hline
 &\multicolumn{2}{c|}{\textbf{$1$-to-$1$}}&\multicolumn{2}{c|}{\textbf{$1$-to-$n$}}&\multicolumn{2}{c|}{\textbf{$n$-to-$1$}}&\multicolumn{2}{c|}{\textbf{$n$-to-$m$}}\\\hline
\textbf{Model}  & \makecell[c]{\textbf{Left}\\ \textbf{\tfhten}} & \makecell[c]{\textbf{Right}\\ \textbf{\tfhten}}& \makecell[c]{\textbf{Left}\\ \textbf{\tfhten}} & \makecell[c]{\textbf{Right}\\ \textbf{\tfhten}}& \makecell[c]{\textbf{Left}\\ \textbf{\tfhten}} & \makecell[c]{\textbf{Right} \\\textbf{\tfhten}}& \makecell[c]{\textbf{Left}\\ \textbf{\tfhten}} & \makecell[c]{\textbf{Right}\\ \textbf{\tfhten}}\\ \hline
\textbf{TransE} & 76.67 & 80.00 & 48.31 & 32.58 & 3.89 & \textbf{78.34} & 63.82 & 80.23\\\hline
\textbf{DistMult} & 83.33 & \underline{83.33} & \underline{49.44} & 29.21 & 6.09 & 55.67 & 60.21 & 79.77\\\hline
\textbf{ComplEx}  & \textbf{90.00} &  \underline{83.33} &\textbf{55.06} & 31.46 & 5.75 & 29.10 & 62.26 & 80.47\\\hline
\textbf{ConvE}    & 83.33 &  80.00 &43.82 & \textbf{39.33}& 3.21 & 74.11 & \underline{65.31} & \textbf{80.98}\\\hline
\textbf{RotatE}   & 83.33  &  \underline{83.33}  & \textbf{55.06} & \underline{38.20} & \textbf{6.43} & \underline{77.66} & 61.66 &  \underline{80.68}\\\hline
\textbf{TuckER}   & \underline{86.67}  &  \textbf{90.00}  & 42.70& \textbf{39.33} & 5.25 & 72.59 & 60.72 &  79.67\\\hline
\textbf{AMIE}     & 73.33  & 73.33  & 13.48 &11.24 & \underline{6.26} & 7.45 & \textbf{67.44} & 64.24 \\ \hline
\end{tabular}
\end{table}


\begin{table}
\caption{\small \fhitone{} results}\label{table:Hits@1}
\scriptsize
\centering
\setlength{\tabcolsep}{2pt}
\renewcommand{\arraystretch}{0.8}
\begin{tabular}{|c|c|c|c|c|} \hline
\textbf{Model}&\textbf{FB15k}&\textbf{FB15k-237}&\textbf{WN18}&\textbf{WN18RR}\\ \hline

\textbf{TransE} &
\makecell[c]{--\\\color{blue}26.9}& 
\makecell[c]{--\\\color{blue}19.1}&
\makecell[c]{--\\\color{blue}31.1}&
\makecell[c]{--\\\color{blue}5.1}\\\hline

\textbf{DistMult}&
\makecell[c]{--\\\color{blue}--\\\color{green}54.1}& 
\makecell[c]{--\\\color{blue}15.5\\\color{green}--}&
\makecell[c]{--\\\color{blue}--\\\color{green}75.2}& 
\makecell[c]{--\\\color{blue}29.1\\\color{green}--}\\\hline

\textbf{ComplEX}&
\makecell[c]{59.9\\\color{blue}--\\\textcolor{green}{59.5}}& 
\makecell[c]{--\\\color{blue}15.9\\\color{green}--}&
\makecell[c]{93.6\\\color{blue}--\\\color{green}93.7}& 
\makecell[c]{--\\\color{blue}34.0\\\color{green}--}\\\hline

\textbf{ConvE}&
\makecell[c]{67.0\\\color{orange}60.7}& 
\makecell[c]{23.9\\\color{orange}23.13}&
\makecell[c]{93.5\\\color{orange}93.9}&
\makecell[c]{39.0\\\color{orange}39.2}\\\hline
\textbf{RotatE}&
\makecell[c]{74.6\\\color{cyan}73.8}& 
\makecell[c]{24.1\\\color{cyan}23.9}&
\makecell[c]{94.4\\\color{cyan}94.4}&
\makecell[c]{42.8\\\color{cyan}42.5}\\\hline
\textbf{TuckER}&
\makecell[c]{74.1\\\color{pink}72.9}& 
\makecell[c]{26.6\\\color{pink}26.2}&
\makecell[c]{94.9\\\color{pink}94.6}&
\makecell[c]{44.3\\\color{pink}42.8}\\\hline
\textbf{AMIE}&
\makecell[c]{--\\\color{velvet}75.1}& 
\makecell[c]{--\\\color{velvet}22.5}&
\makecell[c]{--\\\color{velvet}93.9}&
\makecell[c]{--\\\color{velvet}35.6}\\\hline
\makecell[c]{\textbf{Simple Model} \\(generated by us)}&
\makecell[c]{71.6}& 
\makecell[c]{1.1}&
\makecell[c]{96.4}&
\makecell[c]{34.8}\\\hline
\end{tabular}
\end{table}
\subsection{Results}

Tables~\ref{table:results} and~\ref{table:results wn} display the results of link prediction on FB15k vs.~FB15k-237 and WN18 vs.~WN18RR for all compared methods, using both raw and filtered metrics explained in Section~\ref{sec:eval-measure}.  
For each method, the table shows the original publication where it comes from. The values in black color are the results listed in the original publication, while a hyphen under a measure indicates that the original publication did not list the corresponding value. The values in other colors are obtained through our experiments using various source codes, as listed in the tables.

Below we summarize and explain the results in Table~\ref{table:results} and Table~\ref{table:results wn}. 
(1) The overall observation is that the performance of all methods worsens considerably after removal of reverse relations. For instance, the \fmrr\ of ConvE---one of the best performing methods under many of the metrics---has decreased from 0.698 (on FB15k) to 0.305 (on FB15k-237) and from 0.945 (on WN18) to 0.429 (on WN18RR). Its \fmr\ also became much worse, from 46.5 (FB15k) to 271.5 (FB15k-237) and from 396.6 (WN18) to 4992.7 (WN18RR).  This result verifies that embedding-based methods may only perform well on reverse relations. However, a straightforward approach based on detection of reverse relations can achieve comparable or even better accuracy, as explained in Section~\ref{sec:reverserelation}.

(2) Many successors of TransE (e.g., DistMult, ComplEx, ConvE, RotatE, and TuckER) were supposed to significantly outperform it. 
This was indeed verified by our experiment results on FB15k and WN18. 
However, on FB15k-237, their margin over TransE became much smaller. For example, by \fmrr, TransE's accuracy is 0.288, in comparison with DistMult (0.238), ComplEx (0.249), ConvE (0.305), RotatE (0.337), and TuckER (0.355). We hypothesize that these models improved the results mostly on reverse and duplicate triples and hence, after removing those triples, they do not exhibit clear advantage. This hypothesis can be verified by our finding that most of the test triples on which these models outperformed TransE have reverse or duplicate triples in the training set, as shown in Table~\ref{table:reverse_triple_percentage}. The observations regarding WN18RR are similar, although the successors of TransE demonstrated wider edge over TransE. We note that, however, this could be attributed to the large number of reverse triples from symmetric relations that are retained in WN18RR, as explained in Section~\ref{sec:newdataset}. 

(3) Tables~\ref{table:results} and \ref{table:results wn} show that the performance of AMIE also substantially degenerates in the absence of data redundancy. For example, its \fhitten\ has decreased from 88.1\% (FB15k) to 47.7\% (FB15k-237) and from 94.0\% (WN18) to 35.6\% (WN18RR).

(4) We further analyzed how the most-accurate models perform. 
There are 224, 11, and 34 distinct relations in the test sets of FB15k-237, WN18RR, and YAGO3-10, respectively. Table~\ref{table:num fb} shows, for each metric and each model, the number of distinct test relations on which the model is the most accurate.\footnote{We rounded all accuracy measures to the nearest hundredth except for MRR/FMRR which are rounded to the nearest thousandth. Since there are ties in best performing models, the summation of each column can be greater than 224, 11, and 34.} Furthermore, the heatmap in Figure~\ref{fig:heatmap} (Figure~\ref{fig:heatmap_wn18rr}, resp.) shows, for each of the 224 (11, resp.) relations in FB15k-237 (WN18RR, resp.), the percentage of test triples on which each model has the best performance (i.e., the highest rank) in comparison with other models, using \fmrr{} as the performance measure. 
What is particularly insightful about Figure~\ref{fig:heatmap_wn18rr} is that TuckER, RotatE, and ConvE clearly dominated other models on relations \#1 (\rel{derivationally\_related\_form}), \#8 (\rel{similar\_to}), and \#10 (\rel{verb\_group}). As explained in Section~\ref{sec:newdataset}, these are all symmetric relations where reverse triples are retained in WN18RR. This observation corroborates with the analysis in 2) above. It suggests that the state-of-the-art models might be particularly optimized for reverse triples. On the other hand, the simple rule based on data statistics can attain an \fhitone\ of 97.85\% on these 3 relations.

(5) To better understand the strengths and weaknesses of each model, we further broke down the numbers in the column \fmrr of Table~\ref{table:num fb}. The relations are categorized into 4 different classes: $1$-to-$1$, $1$-to-$n$, $n$-to-$1$ and $n$-to-$m$, based on the average number of heads per tail and tails per head. An average number less than 1.5 is marked as ``$1$'' and ``$n$'' otherwise~\cite{bordes2013translating}.  Among the 224 distinct relations in the test set of FB15k-237, 5.8\% are $1$-to-$1$, 11.6\% are $1$-to-$n$, 33.9\% are $n$-$1$, and 48.7\% are $n$-to-$m$ relations. The numbers of test triples belonging to these 4 types of relations are 192, 1,293, 4,285 and 14,696, respectively. In WN18RR, the 11 distinct relations in the test set are distributed as 2, 4, 3, and 2 in these four classes, and the numbers of test triples are 42, 475, 1,487, and 1,130, respectively.
Figure~\ref{fig:best res cat fb} shows the break-down of relations on which each method has the best result for FB15k-237. 
Figure~\ref{fig:best res cat percentage fb} shows the break-down of best performing models within each type of relations. Overall, RotatE and TuckER outperformed other models, with RotatE particularly excelling on $1$-to-$1$ and $1$-to-$n$ relations and TuckER on $n$-to-$1$ and $n$-to-$m$ relations. TransE still demonstrated its robust strength on $1$-to-$1$ relations.
Our experiment results on WN18RR show some characteristics similar to that of Figure~\ref{fig:res_by_rel_type_fb}. However, since WN18RR has only 11 relations, the distributions are less indicative and robust. Hence, we omit the discussions of such results on WN18RR.

(6) We further computed the \fhitten\ for head and tail predictions separately for each relation type of FB15k-237 and WN18RR, as in Tables~\ref{table:amieh10_fb} and \ref{table:amie_hit10_wn}. The first and second best results are shown with boldface and underline, respectively. All the methods performed better at predicting the ``$1$'' side of $1$-to-$n$ and $n$-to-$1$ relations on both datasets. On FB15k-237, RotatE and TransE are the first and second best performing models on $1$-to-$1$ relations,  respectively. RoataE and TuckER have the highest performance on $1$-to-$n$, $n$-to-$1$, and $n$-to-$m$ relations. Performance of TuckER, RotatE, and AMIE is the best in predicting the side $n$ of $1$-to-$n$ and $n$-to-$1$ relations which are more complicated. On WN18RR, almost all models have very high accuracy on $1$-to-$1$ and $n$-to-$m$ relations. Note that self-reciprocal relations \rel{derivationally\_related\_form}, \rel{similar\_to}, and \rel{verb\_group} belong to these categories.

(7) As mentioned in Section~\ref{sec:duplicate}, YAGO3-10 is dominated by two relations \rel{isAffilatedTo} and \rel{playsFor} which are effectively duplicate relations. The results on this dataset for some the best performing models are shown in Table~\ref{table:results yago}. AMIE achieved better performance than embedding models on \fhitone and \fmrr. Similar to (5) and (6), we generated detailed results on this dataset, shown in Figure~\ref{fig:res_by_rel_type_yago} and Table~\ref{table:amieh10_yago}. Figure~\ref{fig:res_by_rel_type_yago} shows that AMIE outperformed other models on $1$-to-$1$ and $n$-to-$m$ relations while TuckER was on par with AMIE on $1$-to-$1$ relations. RotatE and TuckER outperformed others in $1$-to-$n$ and $n$-to-$1$ relations. Table~\ref{table:amieh10_yago} shows that embedding models have very similar results particularly on $1$-to-$1$ and $n$-to-$m$ relations. We note that the duplicate relations \rel{isAffilatedTo} and \rel{playsFor} belong to $n$-to-$m$ and the self-reciprocal relation \rel{isMarriedTo} belongs to $1$-to-$1$.

(8) The results on YAGO3-10-DR are also shown in Table 11. Same as what we observed on FB15k-237 and WN18RR, the performance of all models dropped significantly after removal of duplicate and reverse triples. Hence, we recommend YAGO3-10-DR for evaluating embedding models instead of YAGO3-10.

(9) Table~\ref{table:Hits@1} compares various methods using \fhitone, which is a more demanding measure than \fhitten, since it only considers whether a model ranks a correct answer at the very top. The results show that embedding models and AMIE have comparable performance on FB15k and WN18 as these datasets contain relations that clearly can be predicted by rules. On FB15k-237 and WN18RR, embedding models RotatE and TuckER stand out.

%% file: sec-conclusions.tex
\section{Conclusions}\label{sec:conclusions}
In this paper, we did an extensive investigation of data redundancy in the widely-used benchmark datasets FB15k, WN18, and YAGO3-10 and its impact on the performance of link prediction models. Our experiments show that, in the absence of the straightforward prediction tasks, the performance of these models degenerates significantly. We identified Cartesian product relations in FB15k which also lead to unrealistic evaluation performance. Given the data characteristics, oftentimes a simple rule based on data statistics can challenge the accuracy of complex machine learning models. Moreover, these problems with the data present unrealistic cases of link prediction that are nonexistent in the real world. We also demonstrated the inadequacy of existing evaluation metrics that penalize a method for generating correct predictions not available in the ground-truth dataset. The results of the study render link prediction a task without truly effective automated solution. Hence, we call for re-investigation of possible effective approaches for completing knowledge graphs.